\documentclass[10pt,twocolumn,letterpaper]{article}

\usepackage{cvpr}      
\usepackage{multirow}
\usepackage{makecell}
\usepackage[table]{xcolor}
\usepackage{pifont}
\usepackage{subcaption}
\usepackage[accsupp]{axessibility}
\newcommand\blfootnote[1]{%
	\begingroup
	\renewcommand\thefootnote{}\footnote{#1}%
	\addtocounter{footnote}{-1}%
	\endgroup
}

\definecolor{cvprblue}{rgb}{0.21,0.49,0.74}
\usepackage[pagebackref,breaklinks,colorlinks,allcolors=cvprblue]{hyperref}

\def\paperID{} 
\def\confName{CVPR}
\def\confYear{2026}

\title{Layer Consistency Matters: Elegant Latent Transition Discrepancy for Generalizable Synthetic Image Detection}

\author{Yawen Yang$^{}$ \quad 
    Feng Li$^{*}$ \quad 
    Shuqi Kong$^{}$ \quad 
    Yunfeng Diao$^{}$ \\
    Xinjian Gao$^{}$ \quad 
    Zenglin Shi$^{}$ \quad 
    Meng Wang$^{}$ \\
    $^{}$Hefei University of Technology
}
{\tt\small }

\begin{document}
\maketitle
\blfootnote{* Corresponding Author}
\begin{abstract}

Recent rapid advancement of generative models has significantly improved the fidelity and accessibility of AI-generated synthetic images. While enabling various innovative applications, the unprecedented realism of these synthetics makes them increasingly indistinguishable from authentic photographs, posing serious security risks, such as media credibility and content manipulation. Although extensive efforts have been dedicated to detecting synthetic images, most existing approaches suffer from poor generalization to unseen data due to their reliance on model-specific artifacts or low-level statistical cues. In this work, we identify a previously unexplored distinction that real images maintain consistent semantic attention and structural coherence in their latent representations, exhibiting more stable feature transitions across network layers, whereas synthetic ones present discernible distinct patterns. Therefore, we propose a novel approach termed latent transition discrepancy (LTD), which captures the inter-layer consistency differences of real and synthetic images. LTD adaptively identifies the most discriminative layers and assesses the transition discrepancies across layers. Benefiting from the proposed inter-layer discriminative modeling, our approach exceeds the base model by 14.35\% in mean Acc across three datasets containing diverse GANs and DMs. Extensive experiments demonstrate that LTD outperforms recent state-of-the-art methods, achieving superior detection accuracy, generalizability, and robustness. The code is available at \url{https://github.com/yywencs/LTD}
\end{abstract}
    
\section{Introduction}
\label{sec:intro}

\begin{figure}[t]
  \centering
  
  \begin{subfigure}[b]{0.48\columnwidth}
    \includegraphics[width=\textwidth]{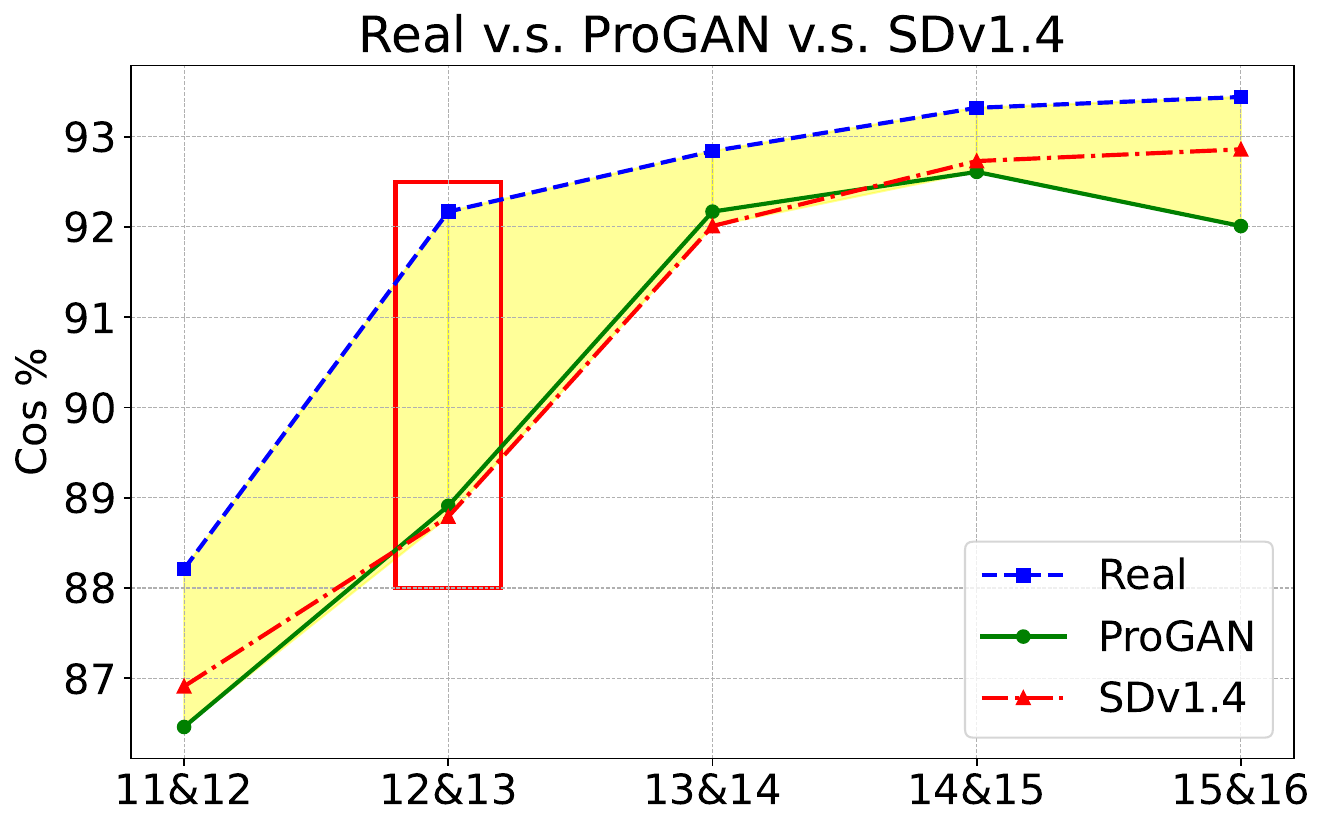}
    \caption{Cosine Similarity}
    \label{fig:sub-a}
    \vspace{-0.3cm}
  \end{subfigure}
  \hfill 
  \begin{subfigure}[b]{0.48\columnwidth}
    \includegraphics[width=\textwidth]{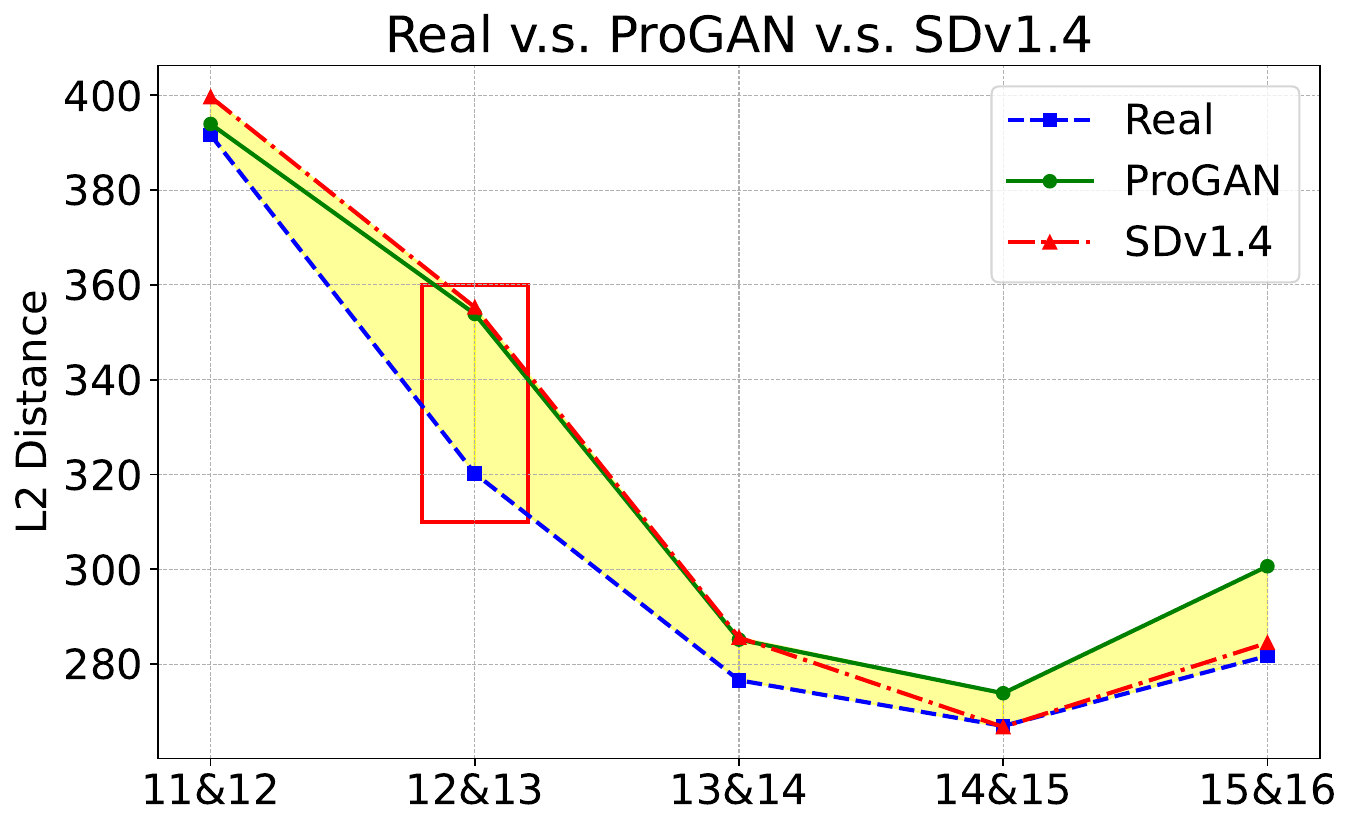}
    \caption{L2 Distance}
    \label{fig:sub-b}
    \vspace{-0.3cm}
  \end{subfigure}
  
  \vspace{3mm} 
  
  \begin{subfigure}[b]{0.48\columnwidth}
    \includegraphics[width=\textwidth]{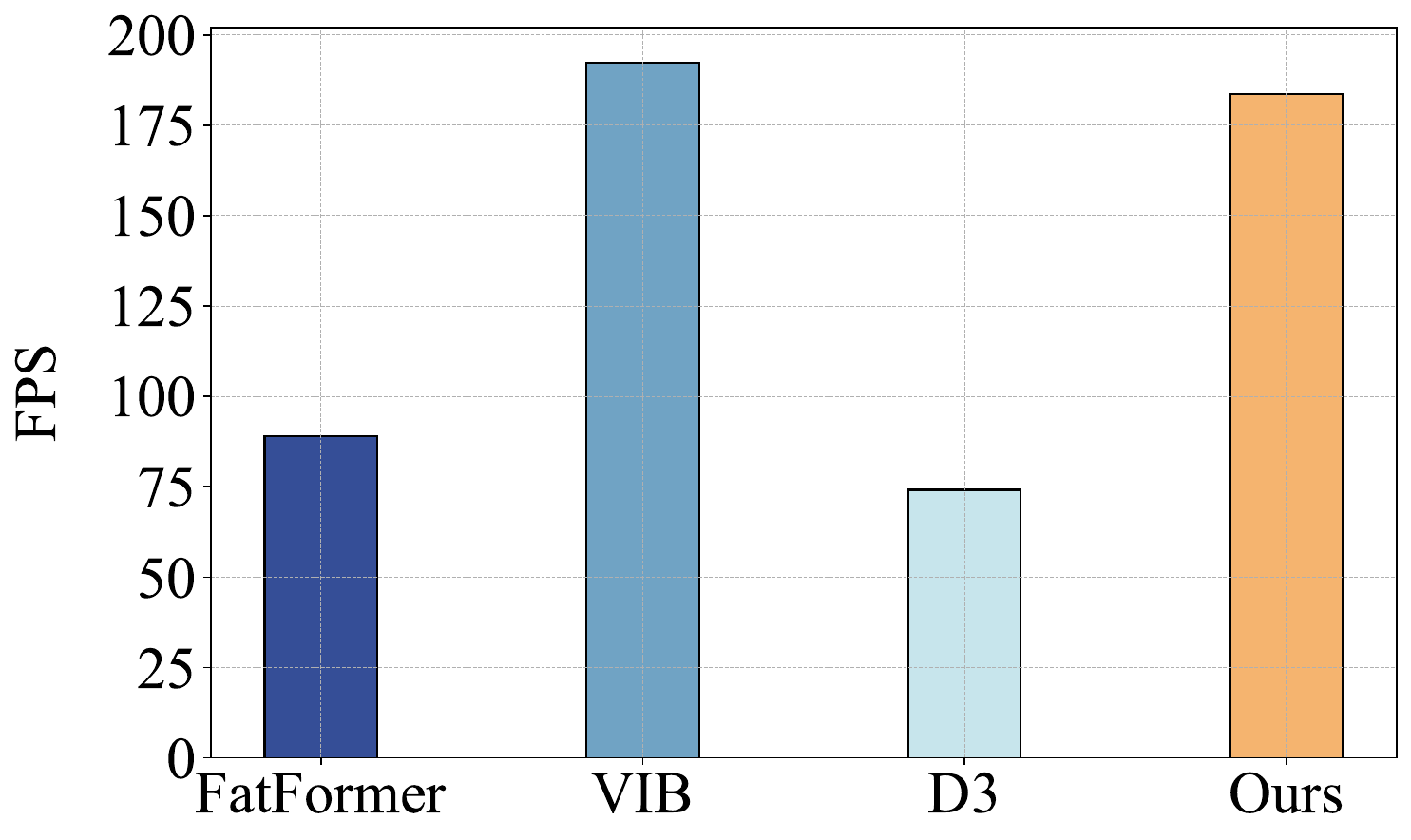}
    \caption{Inference Speed (FPS)}
    \label{fig:sub-c}
  \end{subfigure}
  \hfill 
  \begin{subfigure}[b]{0.48\columnwidth}
    \includegraphics[width=\textwidth]{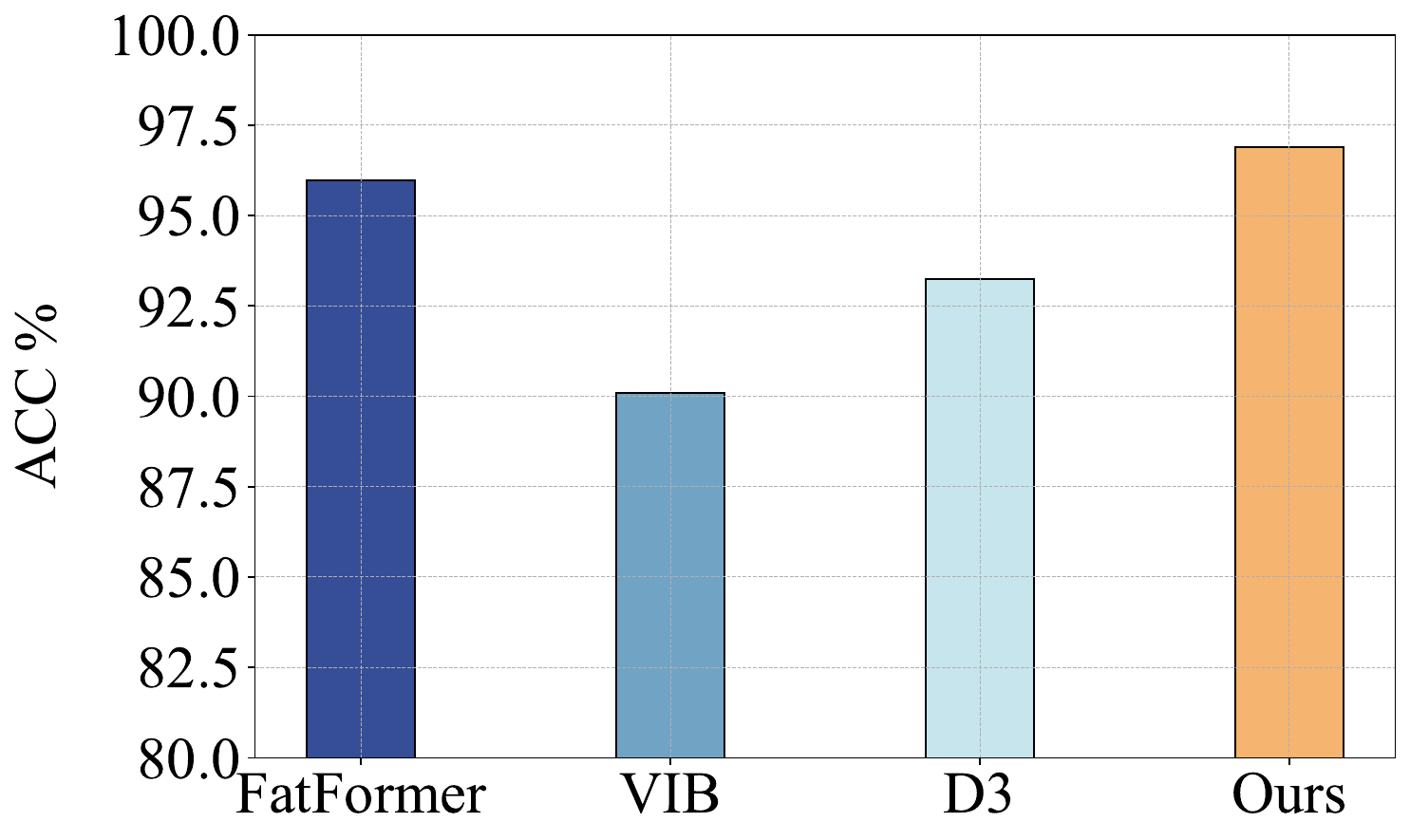}
    \caption{Comparison on UFD}
    \label{fig:sub-d}
  \end{subfigure}
  \vspace{-0.2cm}
  \caption{Illustration of inter-layer feature consistency, computational efficiency, and detection performance. (a) and (b) calculate the cosine similarity and L2 distance between adjacent ViT mid-level layers, respectively. The features from real images maintain stable consistency with lower-layer transition discrepancy (LTD). (c) Inference speed measured in FPS. (d) Detection performance with recent state-of-the-art methods on the UFD~\cite{ojha2023towards} dataset.}
  \label{fig:fig1}
  \vspace{-0.5cm}
\end{figure}
Generative models~\cite{goodfellow2014generative,karras2018progressive,choi2018stargan,brock2018large,park2019semantic,karras2019style,karras2020analyzing} have achieved remarkable progress in synthesizing photorealistic images, a trend greatly accelerated by the advent of diffusion models (DM)~\cite{ho2020denoising,nichol2021glide,dhariwal2021diffusion,gu2022vector,esser2024scaling}, demonstrating impressively realistic and creative expression. However, the dramatic realism and accessibility of synthetic imagery make it increasingly difficult to distinguish from real images, which raises critical concerns about privacy and security due to the potential misuse of deepfakes and misinformation dissemination~\cite{juefei2022countering,feng2025deepfake,qu2025hate}, underscoring the urgent need for robust detectors capable of discerning AI-generated images.  

In response to this challenge, extensive research efforts~\cite {wang2020cnn,liu2022detecting,jeong2022bihpf,jeong2022fingerprintnet,jeong2022frepgan,koutlis2024leveraging,tan2024rethinking} have been directed towards discovering discriminative forensic traces for accurate synthetic image detection. Early studies predominantly address GAN-based generation, where some methods focus on deepfake detection utilizing facial features~\cite{yu2019attributing}, and others target various natural images by capturing generative artifacts in the frequency~\cite{frank2020leveraging,jeong2022bihpf,liu2022detecting,tan2024frequency,jeong2022frepgan} or texture~\cite{wang2020cnn,chai2020makes,tan2023learning,zhong2023patchcraft,tan2024rethinking} domains. While often effective within their training domain, these methods rely on data from a limited set of models (\emph{e.g.}, ProGAN~\cite{karras2018progressive}), causing them to learn model-specific biases that hamper generalization to unseen samples. This limitation is exacerbated by the rise of DMs, whose distinct generation paradigms yield different artifact signatures and a degree of realism that narrows the perceptual gap with real images. Although some subsequent methods~\cite{wang2023dire,luo2024lare,chu2025fire} have been specifically designed to model DM reconstruction errors, they still struggle with the dilemma of cross-domain generalization. 

To address this issue, recent methods~\cite{ojha2023towards,liu2024forgery,koutlis2024leveraging,yang2025d,zhang2025towards,tan2025c2p} have pivoted towards semantic artifacts using pre-trained CLIP model~\cite{radford2021learning}. UnivFD~\cite{ojha2023towards} pioneers a CLIP-based detector that performs nearest neighbor/linear probing on the final CLIP feature for real/fake classification, overlooking the rich low-level information in shallow feature space. Some other methods~\cite{koutlis2024leveraging,liu2024forgery,chen2025forgelens} improve UnivFD by integrating forgery-aware features across all ViT layers~\cite{dosovitskiy2020image}. Despite their effectiveness, these methods actually extract general features that contain substantial irrelevant information. This can introduce noise in learning for forgery-aware traces, thus interfering with detection. 

This work investigates the frozen CLIP-based detection paradigm through the lens of layer transition discrepancies in ViT feature space. By analyzing feature similarity between adjacent layers (Figure~\ref{fig:fig1} (a) and (b)), we observe a notable distinction that, in the mid-level space, real images maintain consistent semantic attention across layers to provide stable feature evolution, whereas synthetic images often show abrupt shifts between foreground and background regions (Figure~\ref{fig:fig2}), thus presenting substantial deviations between consecutive layers. Based on this observation, we propose a novel detection framework that leverages the discriminative power of layer transition discrepancy in frozen CLIP-ViT models. Specifically, our LTD-based detector devises a dynamic layer-wise selection strategy that identifies the most informative subset of consecutive mid-level layers. On these selected candidates, we calculate the differences between adjacent ones to capture the LTD features. Then, a dual-branch architecture is developed, where one branch models holistic feature consistency from the selected raw features, while the other amplifies the LTD. By simultaneously modeling both global structural alignment and local inter-layer variations. Our contributions are as follows:
\begin{itemize}
\item We propose a cross-layer transition representation that leverages the discrepancy in feature evolution across mid-level ViT layers. We provide a comprehensive analysis that real images exhibit more stable layer-wise consistency than synthetic ones, providing a discriminative cue for generalizable synthetic image detection.
\item We propose a dynamic layer-wise selection strategy that adaptively determines the most informative consecutive mid-level layers for each image. Based on this, we build an LTD-based detector that bridges local inter-layer variation and global structural alignment for better detection.

\item Extensive experiments demonstrate the state-of-the-art performance of our method in synthetic image detection, which achieves superior robustness and generalizability across a diverse spectrum of GAN and DM generators with more efficient implementation (Figure~\ref{fig:fig1} (c) and (d)).
\end{itemize}


\section{Related Works}
\label{sec:formatting}
\subsection{Generalizable Synthetic Image Detection}

Generalized detectors focus on analyzing spatial irregularities or specific frequency patterns to distinguish real and fake ones. The spatial line~\cite{marra2018detection,wang2020cnn,tan2023learning,li2025improving,li2025improving,jia2025secret} is dedicated to extracting low-level texture artifacts from images. LNP~\cite{liu2022detecting} discriminates generated images by capturing their inconsistent noise patterns. LGrad~\cite{tan2023learning} uses image gradients as a generalized artifacts representation. 
NPR~\cite{tan2024rethinking} models neighboring pixel relationships to capture upsampling artifacts, achieving source-agnostic detection across both GANs and DMs. There are also several methods~\cite{zhong2023patchcraft,yang2025d} that disrupt the universal artifacts in patches to expose discrepancies in generative patterns. In parallel, the prominent upsampling artifacts that GAN models impart on high-frequency components have motivated the frequency domain~\cite{durall2020watch,frank2020leveraging,luo2021generalizing,jeong2022fingerprintnet}. FreqGAN~\cite{jeong2022frepgan} adversarially learns frequency-level perturbation to suppress domain-specific frequency spectral signatures and accentuates pixel-level irregularities. FingerprintNet~\cite{jeong2022fingerprintnet} simulates generator-specific artifacts by applying varied upsampling rates to real images and performs detection directly in the frequency domain.
BiHPF~\cite{jeong2022bihpf} utilizes bilateral high-pass filters to amplify high-frequency components, particularly within background regions where generative artifacts are more pronounced. FreqNet~\cite{tan2024frequency} leverages both phase and amplitude spectra to strengthen its capability in broader artifact identification. However, these methods mostly emphasize GAN-based forgeries, often ineffective when confronted with diffusion-generated images.
\vspace{-0.3em}
\subsection{Specialized Diffusion Image Detection}
The recent dominance of DMs~\cite{dhariwal2021diffusion,zhang2023adding} in content generation and artistic creation has catalyzed the development of specialized detectors that aim to identify their generated images. Some methods~\cite{wang2023dire,ricker2024aeroblade,chu2025fire} suggest leveraging the reconstruction error based on the observation that synthetic images can be more readily recovered than real counterparts. LaRE$^2$~\cite{luo2024lare} introduces a latent reconstruction error within single-step reconstruction, significantly improving the detection efficiency. DRCT~\cite{chen2024drct} presents the diffusion reconstruction contrastive learning, which generates hard samples by DM and adopts contrastive training to capture diffusion artifacts. FakeInversion~\cite{cazenavette2024fakeinversion} extracts text-conditioned inversion features from Stable Diffusion (SD)~\cite{rombach2022high}, enabling detection of synthetic images from unseen text-to-image generators. FIRE~\cite{chu2025fire} researches the frequency decomposition on the reconstruction error, offering a robust method for DM-generated image detection.

\subsection{Frozen CLIP-based Detector}
Recently, some methods~\cite{ojha2023towards,liu2024forgery,chen2025forgelens} have emerged that look beyond low-level features, focusing instead on semantic-level artifacts by using powerful pre-trained VLMs like CLIP~\cite{radford2021learning}. A straightforward strategy for detection leverages pre-trained embeddings directly, as exemplified by UnivFD~\cite{ojha2023towards}, which performs classification using the frozen final-layer representations of CLIP-ViT. RINE~\cite{koutlis2024leveraging} emphasizes the importance of intermediate representations by combining features from multiple ViT blocks, while other approaches~\cite{liu2024forgery,chen2025forgelens} insert lightweight, forgery-aware learners between adjacent ViT blocks to enhance sensitivity to forgery cues. On the textual side, C2P-CLIP~\cite{tan2025c2p} enhances semantic alignment by integrating a category-common prompt into the CLIP text encoder and contrastively constrains the classification through concept matching. VIB~\cite{zhang2025towards} imposes an information bottleneck on CLIP-ViT features to suppress redundant forgery-invariant information. This work argues that modeling the layer transition discrepancy of images is crucial for synthetic image detection. Our method innovatively bridges local inter-layer variation and global structural alignment, thus demonstrating superior detection accuracy across diverse generators.

\section{Methods}
\subsection{Motivation}
\label{subsec:analysis}
Existing CLIP-based detectors follow a common paradigm that applies pre-trained visual encoders (mostly ViT-L/14) to extract general-purpose representations. They typically exploit the rich semantic information from the final embeddings~\cite{ojha2023towards,zhang2025towards} or all intermediate features~\cite{liu2024forgery,koutlis2024leveraging,chen2025forgelens} to facilitate the classification of real/fake images. Given an input image $\mathbf{x}$, the CLIP-ViT encoder first partitions it into a sequence of non-overlapping patches. These patches are then processed through a stack of transformer layers. This layered architecture induces a natural hierarchy of visual representations: the initial layers capture low-level statistics, such as fine-grained textures; mid-level layers integrate these local cues into structured representations that reflect object parts and scene layouts; while deeper layers encode semantic concepts~\cite{dosovitskiy2020image, raghu2021vision}.

\begin{figure}[t]
    \centering
    \includegraphics[width=\columnwidth]{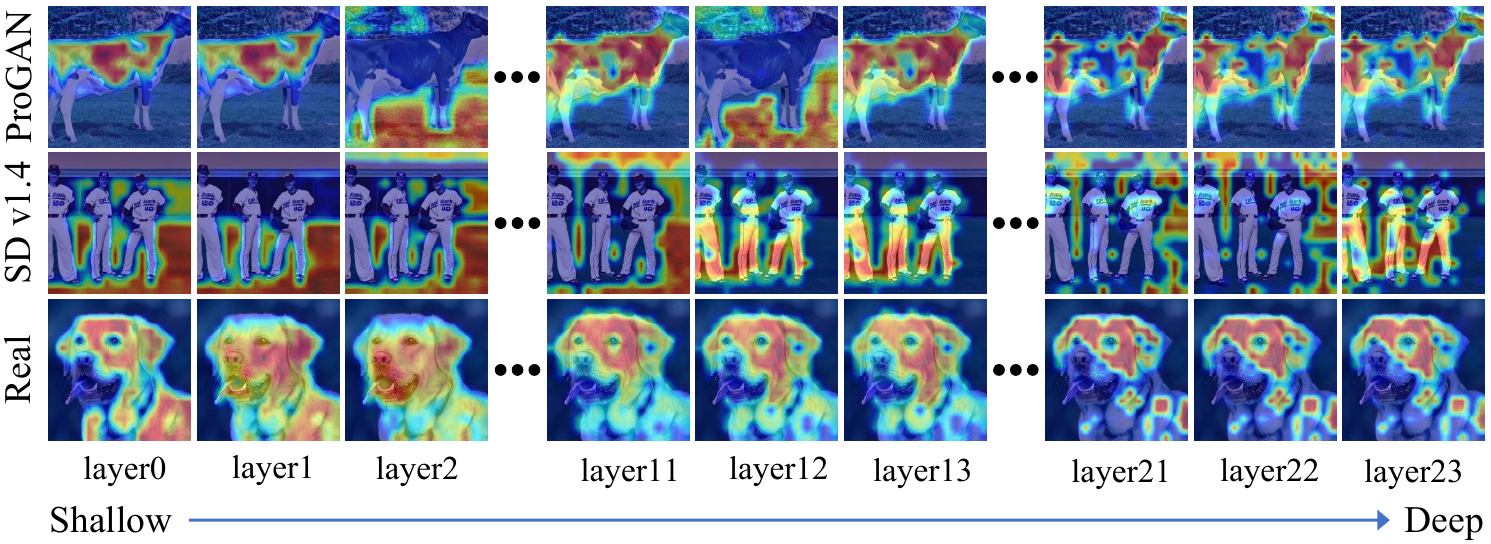} 
    \vspace{-0.5cm}
    \caption{CAM visualization of features extracted by the CLIP ViT-L/14 transformer layers for ProGAN, SD v1.4, and real images. ProGAN and SD v1.4 show noticeable attention shifts between foreground and background regions in mid-level features, indicating unstable representations and discrepancies in layer transition. Real images exhibit highly consistent attention across adjacent layers, reflecting smooth and stable evolution.}
    \label{fig:fig2}
    \vspace{-1em}
\end{figure}

However, in this work, we revisit the CLIP-encoded features and shift focus to the inter-layer transition discrepancies that have been overlooked in the current research community. To this end, we analyze spatial attention dynamics across hierarchical layers using class activation maps (CAMs). As illustrated in Figure~\ref{fig:fig2}, we can observe that: 1) In shallow layers (\emph{e.g.}, Layer0 \emph{v.s.} Layer1), both real and synthetic images exhibit substantial embedding space overlap, indicating low separability. This is consistent with the known function of the early network stage, which primarily captures low-level visual primitives sharing similar perceptual properties between real and synthetic images; 2) Deep layers (\emph{e.g.}, Layer22 \emph{v.s.} Layer23) also demonstrate limited separation as they prioritize semantic coherence over structural details, both image types converge toward similar semantic manifolds due to the text-image alignment objective of CLIP, thus reducing discriminability; 3) Surprisingly, intermediate layer transitions (\emph{e.g.}, Layer11 \emph{v.s.} Layer12 \emph{v.s.} Layer13) show distinct separation patterns, suggesting that mid-level features possess unique potential for capturing discriminative clustering patterns that effectively distinguish real from synthetic images. 

To intuitively investigate the inter-layer transition discrepancy of two types of images, we conduct a layer-wise differential analysis using t-SNE visualization on ProGAN and SD v1.4, as shown in Figure~\ref{fig:fig3}. It reveals that the most discriminative discrepancies for detecting synthetic content emerge specifically during mid-level layer transitions. We attribute this to the inherent optimization paradigm of modern generative pipelines: while prioritizing pixel-level realism and high-level semantic alignment, they often lack strict physical constraints. Consequently, as mid-level ViT layers integrate local textures into structural components, synthetic images fail to maintain continuous spatial correlations, thereby exposing abrupt structural artifacts. This phenomenon appears to reflect an intrinsic property of the generative pipeline itself, which exploits a model-agnostic signature embedded in the hierarchical feature evolution, making our approach fundamentally different from existing methods with improved robustness and generalization.

\begin{figure*}[t]
  \centering
    \includegraphics[width=\textwidth]{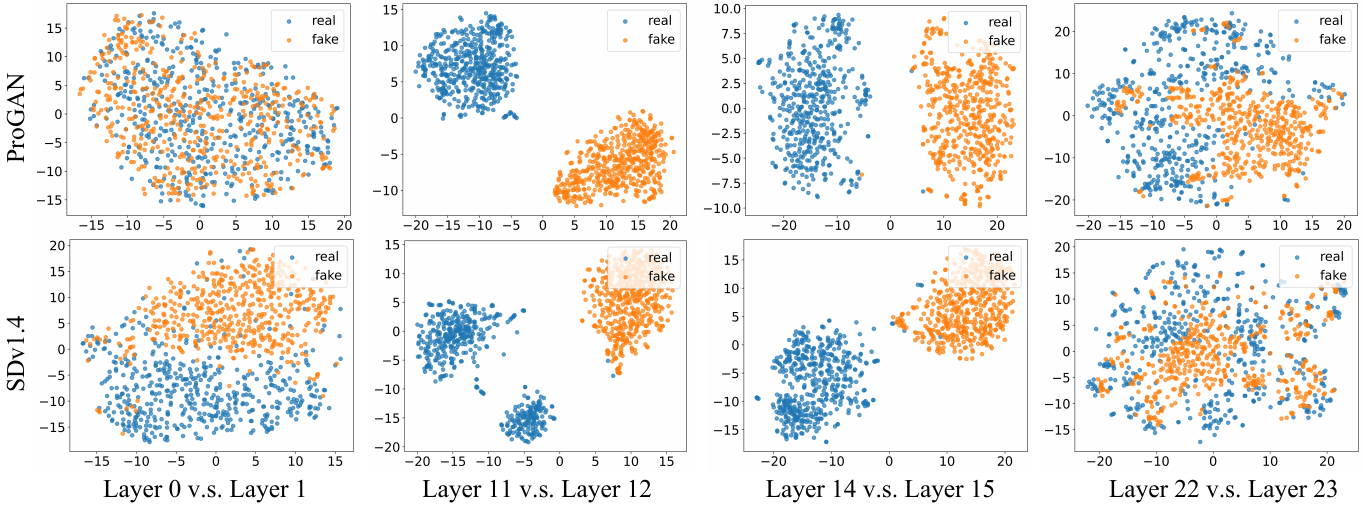} 
    \vspace{-0.6cm}
  \caption{t-SNE maps of the layer transition discrepancy (LTD) between adjacent layers for real and generated images. We can see that \textbf{1)} Shallow layers (Layer 0 vs. 1) and deep layers (Layer 22 \emph{v.s.} 23) show high consistency in both real and generated images, offering limited discriminative power; \textbf{2)} Middle layers (Layer 10 vs. 11 and Layer 14 vs. 15) exhibit a clearer gap of the LTD between real and fake images, providing more discriminative clues for detection.}
  \label{fig:fig3}
  \vspace{-0.2cm}
\end{figure*}

\begin{figure*}[t]
    \centering
    \includegraphics[width=\textwidth]{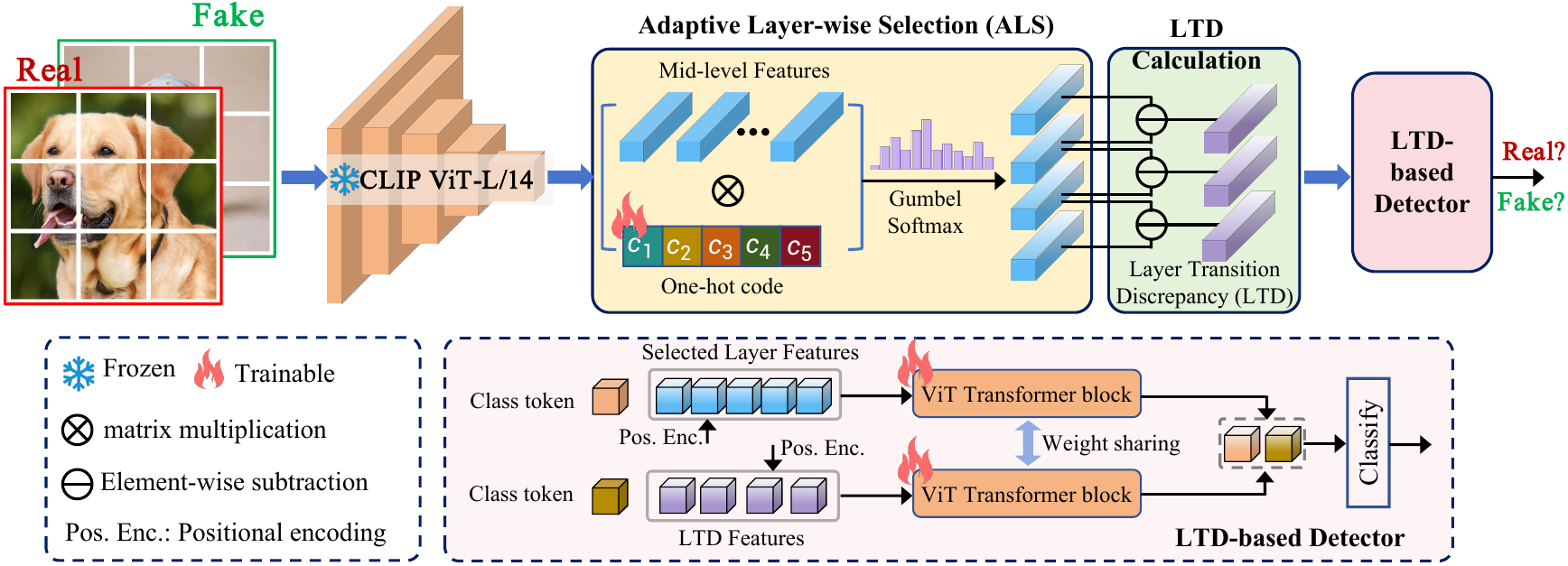} 
    \vspace{-0.5cm}
    \caption{Overview of the proposed Layer Transition Discrepancy(LTD) architecture. Discriminative intermediate features are extracted from a pre-trained Vision Transformer and used to construct two branches: one stacking multi-layer features directly, and the other computing pairwise differences between adjacent layers. A shared transformer block processes both branches, followed by an MLP for classification based on the concatenated representation.}
    \vspace{-0.4cm}
    \label{fig:fig4}
\end{figure*}

\setlength{\tabcolsep}{3pt}
\begin{table*}[t]
\centering
\caption{Performance comparison on UFD in Acc. (\%). Each row corresponds to a detection method, and each column to a generative model. The last column reports the mean Acc across all models. Red and blue colors indicate the best and second-best results, respectively.}
\resizebox{\textwidth}{!}{
\begin{tabular}{lccccccccccccccccc}
\toprule
\multirow{2}{*}{Method} & \multicolumn{7}{c}{GAN-based Models} & \multirow{2}{*}{\makecell{Deep-\\Fake}} & \multicolumn{3}{c}{LDM} & \multirow{2}{*}{ADM} & \multicolumn{3}{c}{Glide} & \multirow{2}{*}{Dalle} & \multirow{2}{*}{\makecell{\textbf{Mean} \\\textbf{Acc}}} \\
\cmidrule(l){2-8} \cmidrule(l){10-12} \cmidrule(l){14-16}
& \makecell{Pro-\\GAN} & \makecell{Style-\\GAN} & \makecell{Style2-\\GAN} & \makecell{Big-\\GAN} & \makecell{Cycle-\\GAN} & \makecell{Star-\\GAN} & \makecell{Gau-\\GAN} &  & 100 & 200 & 200\_cfg & & 50\_27 & 100\_10 & 100\_27 &  \\
\midrule
CNNSpot~\cite{wang2020cnn} & {100.00} & 85.72 & 83.39 & 70.17 & 85.71 & 91.70 & 78.93 & 53.79 & 55.15 & 53.90 & 53.90 & 61.00 & 62.75 & 61.00 & 60.35 & 56.10 & 69.60 \\
UnivFD~\cite{ojha2023towards} & 99.81 & 84.93 & 74.96 & 95.08 & 98.34 & 95.75 & {99.47} & 68.51 & 95.00 & 94.40 & 74.00 & 69.65 & 79.05 & 77.90 & 78.50 & 87.30 & 85.79 \\
LGrad~\cite{tan2023learning} & 99.83 & 94.73 & 96.05 & 85.62 & 84.96 & 99.62 & 72.49 & 56.48 & 93.20 & 92.60 & 94.05 & 77.50 & 89.05 & 88.10 & 88.10 & 86.65 & 87.44 \\
NPR~\cite{tan2024rethinking} & 99.79 & 97.71 & 98.39 & 84.35 & 96.84 & 99.35 & 82.50 & 80.22 & 98.65 & 98.45 & 98.30 & 75.15 & 97.55 & 97.60 & 97.10 & 93.85 & 93.49 \\
FatFormer~\cite{liu2024forgery} & 99.70 & 97.13 & 98.70 & 99.50 & 99.36 & 99.70 & 99.30 & 93.27 & 98.50 & 98.50 & 94.80 & 75.92 & 94.31 & 94.15 & 94.13 & 98.70 & \color{blue}{\underline{95.98}}  \\
RINE~\cite{liu2022detecting} & 100.00 & 88.90 & 94.5& 99.60 & 99.30 & 99.50 & 99.80 & 80.60 & 98.60 & 98.30 & 88.20 & 76.10 & 92.60 & 90.70 & 88.90 & 95.00 & 93.16 \\
FreqNet~\cite{tan2024frequency} & 99.60 & 90.20 & 87.9& 90.50 & 95.10 & 85.70 & 93.40 & 88.90 & 97.90 & 97.50 & 97.40 & 67.30 & 86.70 & 87.90 & 84.50 & 97.40 & 90.49\\
C2P-CLIP~\cite{tan2025c2p} & 99.98 & 96.44 & 93.48& 99.12& 97.31& 99.6& 99.17 & 93.77 & 99.30 & 99.25 & 97.25 & 69.00 & 95.25 & 95.25 & 96.10 & 98.55 & 95.55 \\
$\mathrm{D^{3}}$~\cite{yang2025d} & 99.39 & 94.88 & {99.59} & {99.10} & 90.66 & 95.87 & 98.02 & 67.92 & 94.45 & 94.70 & 88.30 & {93.45} & {94.55} & 94.60 & 94.30 & 92.25 & 93.25 \\
VIB~\cite{zhang2025towards} & 99.90 & 90.45 & 87.25 & 95.20 & 98.35 & 97.25 & 99.25 & 81.70 & 95.70 & 77.70 & 96.25 & 72.70 & 86.25 & 89.25 & 86.95 & 87.40 & 90.10 \\
ForgeLens~\cite{chen2025forgelens} & 99.94 & 97.33 & 97.38& 98.02 & 98.79 & 92.47 & 98.72 & 88.91 & 98.70 & 98.80 & 98.15 & 79.45 & 97.40 & 96.15 & 97.85 & 97.95 & 95.56  \\
\hline
\rowcolor{lightgray}\textbf{LTD (ours)} & 99.98 & 99.10 & 95.70 & 95.53 & 98.00 & 99.50 & 95.38 & 95.20 & 97.00 & 98.50 & 98.00 & 88.00 & 97.00 & 97.50 & 97.50 & 98.50 & \color{red}{\textbf{96.90}} \\
\bottomrule
\end{tabular}
}
\label{tab:tab1}
\end{table*}
\setlength{\tabcolsep}{3pt}
\begin{table*}[t]
\centering
\caption{Performance comparison on UFD in AP. Each row corresponds to a detection method, and each column to a generative model. The last column reports the mean AP across all models. Red and blue colors indicate the best and second-best results, respectively.}

\resizebox{\textwidth}{!}{
\begin{tabular}{lcccccccccccccccccc}
\toprule
\multirow{3}{*}{Method} & \multicolumn{7}{c}{GAN-based Models} & \multirow{2}{*}{\makecell{Deep-\\Fake}} & \multicolumn{3}{c}{LDM} & \multirow{2}{*}{ADM} & \multicolumn{3}{c}{GLIDE} & \multirow{3}{*}{DALL-E} & \multirow{2}{*}{\makecell{\textbf{Mean}\\\textbf{AP}}} \\
\cmidrule(l){2-8} \cmidrule(l){10-12} \cmidrule(l){14-16}
& \makecell{Pro-\\GAN} & \makecell{Style-\\GAN} & \makecell{Style2-\\GAN} & \makecell{Big-\\GAN} & \makecell{Cycle-\\GAN} & \makecell{Star-\\GAN} & \makecell{Gau-\\GAN} &  & 100 & 200 & 200\_cfg & & 50\_27 & 100\_10 & 100\_27 &  \\
\midrule
CNNSpot~\cite{wang2020cnn} & 100.00 & 99.54 & 99.06 & 84.51 & 93.43 & 98.15 & 89.49 & 89.02 & 72.50 & 71.11 & 71.11 & 73.96 & 84.6 & 82.01 & 80.47 & 71.27 & 85.01 \\
UnivFD~\cite{ojha2023towards} & 100.00 & 97.56 & 97.90 & 99.27 & 99.80 & 99.37 & 99.98 & 81.79 & 99.27 & 99.32 & 92.50 & 87.64 & 95.56 & 94.96 & 95.28 & 97.47 & 96.10\\
LGrad~\cite{tan2023learning} & 100.00 & 99.86 & 99.89 & 90.73 & 93.61 & 99.98 & 79.28 & 72.67 & 98.36 & 98.18 & 98.57 & 87.06 & 94.53 & 94.20 & 93.23 & 95.72 & 93.49 \\
NPR~\cite{tan2024rethinking} & 99.99 & 99.78 & 99.94 & 87.80 & 98.94 & 99.94 & 85.49 & 82.40 & 99.80 & 99.76 & 99.73 & 80.29 & 99.61 & 99.63 & 99.55 & 98.96 & 95.73 \\
FatFormer & 100.00 & 99.75 & 99.92 & 99.98 & 99.99 & 100.00 & 100.00 & 97.99 & 99.89& 99.83& 99.22& 91.92& 99.50 & 99.33& 99.27& 99.84& 99.15 \\
RINE~\cite{liu2022detecting} & 100.00 & 99.40 & 100.00 & 99.90 & 100.00 & 100.00 & 100.00 & 97.90 & 99.90 & 99.90 & 98.00 & 95.70 & 99.30 & 98.80 & 98.90 & 93.00 & 98.79 \\
FreqNet~\cite{tan2024frequency} & 100.00 & 99.70 & 99.50 & 96.00 & 99.50 & 99.80 & 98.60 & 94.40 & 99.90 & 99.90 & 99.90 & 75.70 & 96.30 & 96.40 & 96.00 & 99.80 & 96.96\\
C2P-CLIP~\cite{tan2025c2p} & 100.00 & 99.50 & 98.56 & 99.96 & 100.00 & 100.00 & 100.00 & 98.59 & 99.98 & 99.99 & 99.83 & 94.13 & 99.32 & 99.38 & 99.30 & 99.94 & \color{blue}{\underline{99.28}} \\
$\mathrm{D^{3}}$~\cite{yang2025d} & 100.00 & 99.18 & 99.38 & 99.97 & 97.69 & 99.36 & 99.94 & 87.09 & 99.54 & 99.51 & 95.88 & 98.67 & 99.05 & 98.89 & 98.91 & 98.26 & 98.21\\
VIB~\cite{zhang2025towards} & 100.00 & 98.73 & 97.23 & 98.98 & 99.80 & 99.54 & 99.88 & 92.05 & 99.37 & 99.42 & 92.66 & 87.09 & 98.04 & 97.67 & 97.47 & 96.81 & 97.17 \\
ForgeLens~\cite{chen2025forgelens} & 100.00 & 99.85 & 99.58 & 99.99 & 100.00 & 100.00 & 100.00 & 96.23 & 99.96 & 99.95 & 99.72 & 94.82& 99.48 & 99.43 & 99.60 & 99.89 & 99.11 \\
\hline
\rowcolor{lightgray}\textbf{LTD (ours)} & 100.00 & 99.96 & 99.31 & 99.97 & 100.00 & 100.00 & 99.92 & 98.44 & 99.48 & 99.46 & 99.59 & 96.83 & 99.73 & 99.75 & 99.71 & 99.97 & \color{red}{\textbf{99.51}}  \\
\bottomrule
\end{tabular}
                                                                                                                                                                                                                                                                                                                                                                                                                                                                                                                                                                                                                                                                                                                                                                                                                                                                                                                                                                                                                                                                                                                                                                                                                                                                                                                                                                                                                                                                                                                                                                                                                                                    }
\label{tab:tab2}
\end{table*}

\subsection{Detector with Layer Transition Discrepancy}
Motivated by the above analysis, this work designs a simple yet effective detection network based on the layer transition discrepancy (LTD), as illustrated in Figure~\ref{fig:fig4}. Specifically, we utilize the CLIP ViT-L/14 image encoder as the frozen backbone to extract hierarchical visual representations. According to the findings in Sec.~\ref{subsec:analysis} which reveal that the most distinctive discrepancies between real and generated images emerge in the mid-to-high layers, we can directly determine a sequence of intermediate layers, $\mathbf{f}^{(1)}_{mid}, \mathbf{f}^{(2)}_{mid}, ..., \mathbf{f}^{(l)}_{mid}$, to construct discriminative representations.

\textbf{Layer Transition Discrepancy Calculation}. Instead of relying on manually fixed layer combinations, a dynamic layer-wise selection strategy is introduced that adaptively identifies the most informative subset of layers for each input. Let $n$ denote the window size, a fixed hyperparameter specifying the number of consecutive layers to be selected. We consider all possible windows of $n$ consecutive layers as candidates. Then we parameterize the selection with learnable logits $\boldsymbol{\pi} \in \mathbb{R}^C$ (where $C = l - n + 1$) and determine the optimal starting index $s$ via the Gumbel-Softmax to maintain differentiability during training:
\begin{equation}
s = \text{one\_hot}\left(\arg\max_{i} (\log \pi_i + g_i) / \tau\right),
\end{equation}
where $g_i$ denotes Gumbel noise and $\tau$ is the temperature parameter. This enables the differentiable selection of $n$ consecutive layers $\{\mathbf{f}^{(k)}_s\}_{k=1}^n$ from the candidate windows, where each $\mathbf{f}$ denotes the class token (``CLS'') of the corresponding layer. We then compute the LTD by taking feature differences between adjacent selected layers:
\begin{equation}
   \mathbf{d}^{(k)}_s = \mathbf{f}^{(k+1)}_{s} - \mathbf{f}^{(k)}_{s}, \quad k \in \{1, \dots, n-1\}.
\end{equation}
Compared to raw mid-level features, LTD focuses on characterizing inter-layer variations while suppressing redundant irrelevant information.

\textbf{LTD-based Detector}. We design a dual-branch detection architecture in which each branch serves a complementary role. One processes the selected raw features to model holistic feature consistency, and the other amplifies local transition patterns between adjacent layers. Following the standard ViT paradigm, we prepend ``CLS'' $\mathbf{f}_{\text{cls}}$ and $\mathbf{d}_{\text{cls}}$ to the sequence. Considering that these features are extracted from different transformer layers, we add the learnable positional encoding to ensure position awareness in relation to their original hierarchy, constituting the overall features as
\begin{equation}
    \mathbf{F}_s = [\mathbf{f}_{s}, \mathbf{f}_{\text{cls}}, \mathbf{f}_p], \quad \mathbf{D} = [\mathbf{d}, \mathbf{d}_{\text{cls}}, \mathbf{d}_p]
\end{equation}
where $\mathbf{f}_p$ and $\mathbf{d}_p$ denote the positional embeddings for the selected raw layer and LTD features, respectively. The outputs from both branches are processed by parallel, weight-shared trainable ViT transformer blocks for interactive feature learning. This weight-sharing mechanism enforces feature alignment by mapping both spatial consistency and inter-layer transitions into a unified semantic space. This strategy effectively prevents distributional divergence and encourages the model to capture discriminative variations across layers. The transformed representations are concatenated and fed into a classification head for detection.

In this way, our model dynamically selects the most informative depth region and models hierarchical correlations among transformer layers, which effectively bridges local inter-layer variation and global structural alignment, achieving preferable robustness and generalizability across diverse DM and GAN generation methods.

\section{Experiment}

\begin{table*}[t]
\scriptsize
\centering
\caption{Performance comparison on DRCT-2M in Acc. (\%). Each row corresponds to a detection method, and each column to a generative model. The last column reports the mean Acc across all models. Red and blue colors indicate the best and second-best results, respectively.}
\resizebox{\textwidth}{!}{
\begin{tabular}{lcccccccccccccc}
\toprule
\multirow{3}{*}{Method} & \multicolumn{6}{c}{SD variants} & \multicolumn{2}{c}{Turbo Variants} & \multicolumn{2}{c}{LCM Variants} & \multicolumn{3}{c}{ControlNet Variants} & \multirow{2}{*}{\makecell{{Mean}\\{Acc}}} \\
\cmidrule(l){2-7} \cmidrule(l){8-9} \cmidrule(l){10-11} \cmidrule(l){12-14}
& LDM & SDv1.4 & SDv1.5 & SDv2 & SDXL & \makecell{SDXL-\\Refiner} & \makecell{SD-\\Turbo} & \makecell{SDXL-\\Tubo} & \makecell{LCM-\\SDv1.5} & \makecell{LCM-\\SDXL} & \makecell{SDv1-\\Ctrl} & \makecell{SDv2-\\Ctrl} & \makecell{SDXL-\\Ctrl}  \\
\midrule
CNNSpot~\cite{wang2020cnn} & 74.26 & 99.87 & 99.88 & 87.44 & 54.04 & 54.48 & 84.00 & 80.04 & 98.59 & 64.10 & 79.46 & 63.11 & 68.21 & 77.50 \\
UnivFD~\cite{ojha2023towards} & 97.93 & 95.40 & 93.73 & 93.73 & 89.25 & 91.47 & 86.20 & 84.73 & 94.64 & 89.87 & 88.71 & 80.86 & 74.89 & 89.34  \\
LGrad~\cite{tan2023learning} & 52.07 & 51.46 & 51.49& 50.38 & 50.16 & 50.27 & 50.07 & 50.91 & 50.24 & 49.99 & 50.10 & 50.47 & 51.77 & 50.72  \\
NPR~\cite{tan2024rethinking} & 50.31 & 99.70 & {{99.65}} & 54.51 & 50.01 & 50.17 & {{99.30}} & 99.61 & {{99.92}} & 50.17 & 61.47 & 53.99 & 64.76 & 71.81  \\
DIRE~\cite{wang2023dire} & 57.43 & 98.95 & 98.85 & 65.87 & 59.63 & 63.95 & 89.48 & 88.00 & 97.42 & 70.35 & 90.28 & 85.67 & 82.67 & 80.66\\
DRCT~\cite{chen2024drct} & {{99.58}} & 98.64 & 98.48 & {{99.85}} & 96.40 & 97.88 & {{99.42}} & 83.52 & 98.45 & 93.64 & 96.17 & {{99.67}} & 97.48 & 96.86 \\
AIDE~\cite{yan2024sanity} & 80.20 & 62.05 & 62.99 & 71.07 & 50.21 & 58.45 & 53.72 & 52.10 & 52.37 & 50.02 & 53.95 & 52.49 & 61.19 & 58.52 \\
$\mathrm{D^{3}}$~\cite{yang2025d} & 79.73 & 98.91 & 98.91 & 97.38 & 82.39 & 81.24 & 82.35 & 82.35 & 97.40 & 85.75 & 93.70 & 87.12 & 82.08 & 88.41 \\
VIB~\cite{zhang2025towards} & 65.06 & 85.09 & 85.32 & 82.56 & 73.78 & 80.60 & 75.67 & 70.23 & 78.70 & 68.54 & 81.80 & 76.88 & 65.70 & 76.15 \\
ForgeLens~\cite{chen2025forgelens} & 98.56 & {{100.00}} & 98.46 & 98.09 &{{97.84}} & {{99.00}} & 98.10 & 97.58 & 98.01 & {{97.97}} & {{97.20}} & 97.48 & {{98.56}} & \color{blue}{\underline{98.22}} \\
\hline
\rowcolor{lightgray}{LTD (ours)} & {{99.80}} & {{99.84}} & {{99.84}} & {{98.56}} & {{99.65}} & {{99.69}} & {{99.84}} & {{99.84}} & {{99.84}} & {{99.80}} & {{99.56}} & {{98.40}} & {{99.34}} & \color{red}{\textbf{99.54}} \\
\bottomrule
\end{tabular}
}
\label{tab:tab3}
\end{table*}

\begin{table*}[t]
\scriptsize
\centering
\caption{Performance comparison on DRCT-2M in AP. Each row corresponds to a detection method, and each column to a generative model. The last column reports the mean AP across all models. Red and blue colors indicate the best and second-best results, respectively.}
\resizebox{\textwidth}{!}{
\begin{tabular}{lcccccccccccccc}
\toprule
\multirow{3}{*}{Method} & \multicolumn{6}{c}{SD variants} & \multicolumn{2}{c}{Turbo Variants} & \multicolumn{2}{c}{LCM Variants} & \multicolumn{3}{c}{ControlNet Variants} & \multirow{2}{*}{\makecell{{Mean}\\{AP}}} \\
\cmidrule(l){2-7} \cmidrule(l){8-9} \cmidrule(l){10-11} \cmidrule(l){12-14}
& LDM & SDv1.4 & SDv1.5 & SDv2 & SDXL & \makecell{SDXL-\\Refiner} & \makecell{SD-\\Turbo} & \makecell{SDXL-\\Tubo} & \makecell{LCM-\\SDv1.5} & \makecell{LCM-\\SDXL} & \makecell{SDv1-\\Ctrl} & \makecell{SDv2-\\Ctrl} & \makecell{SDXL-\\Ctrl}  \\
\midrule
CNNSpot~\cite{wang2020cnn} & 91.53 & {{100.00}} & {{100.00}} & 99.68 & 92.87 & 92.34 & 99.47 & 98.94 & 99.97 & 87.39 & 92.37 & 97.12 & 86.63 & 95.25 \\
UnivFD~\cite{ojha2023towards}  & 99.89 & 99.72 & 99.72 & 99.62 & 99.28 & 99.55 & 98.96 & 98.80 & 99.67 & 99.26 & 99.24 & 98.25 & 97.53 & 99.19 \\
LGrad~\cite{tan2023learning} & 52.63 & 60.23 & 59.36 & 57.18 & 44.30 & 51.36 & 50.30 & 46.50 & 49.86 & 39.04 & 49.61 & 51.94 & 54.58 & 51.30 \\
NPR~\cite{tan2024rethinking} & 99.60 & {{99.98}} & {{99.98}} & 97.17 & 70.99 & 76.71 & {{100.00}} & {{100.00}} & {{100.00}} & 84.86 & 99.05 & 97.81 & 99.29 & 94.26  \\
DIRE~\cite{wang2023dire} & 90.43 & 99.96 & 99.96 & 93.13 & 92.63 & 95.16 & 99.45 & 99.28 & 99.89 & 96.68 & 99.33 & 99.21 & 98.97 & 97.24 \\
DRCT~\cite{chen2024drct} & 99.99 & 99.86 & 99.86 & {{99.98}} & {{99.98}} & 99.05 & {{100.00}} & 98.12 & {{99.98}} & 98.25 & 99.67 & {{100.00}} & 99.62 & 99.57 \\
AIDE~\cite{yan2024sanity}  & 96.78 & 87.40 & 87.79 & 92.76 & 68.47 & 90.81 & 80.38 & 74.04& 71.61 & 77.00 & 72.17 & 72.54 & 91.21 & 81.77 \\
$\mathrm{D^{3}}$~\cite{yang2025d}  & 99.55 & {{99.98}} & {{99.98}} & 97.38 & 99.01 & 99.45 & 82.35 & 99.22 & 99.94 & {{99.37}} & 99.86 & 99.62 & 99.38 & 98.08 \\
VIB~\cite{zhang2025towards} & 93.21 & 98.39 & 98.38 & 98.11 & 97.10 & 98.10 & 96.93 & 95.77 & 97.02 & 95.61 & 97.69 & 97.00 & 93.90 & 96.71 \\
ForgeLens~\cite{chen2025forgelens} & {{100.00}} & {{100.00}} & {{100.00}} & 99.87 & {{99.57}} & {{99.71}} & {{99.99}} & {{99.95}} & 98.01 & 98.88 & {{99.98}} & 99.10 & {{99.75}} & \color{blue}{\underline{99.76}} \\
\hline
\rowcolor{lightgray}{LTD (ours)} & {{99.99}} & {{100.00}} & {{100.00}} & {{99.94}} & {{99.98}} & {{99.99}} & {{100.00}} & {{100.00}} & {{100.00}} & {{100.00}} & {{99.99}} & {{99.95}} & {{99.98}} & \color{red}{\textbf{99.99}}  \\
\bottomrule
\end{tabular}
}
\vspace{-0.3em}
\label{tab:tab4}
\end{table*}

\subsection{Settings}
\noindent\textbf{Training Dataset}. We evaluate our method on three benchmarks: \textbf{UFD}~\cite{ojha2023towards}, \textbf{DRCT-2M}~\cite{chen2024drct}, and \textbf{GenImage}~\cite{zhu2023genimage}. For UFD and cross-dataset, we adopt a 2-class training setting (\textit{chair}, \textit{tvmonitor}) comprising 72k ProGAN-generated images and real images from LSUN~\cite{yu2015lsun}.To evaluate cross-dataset generalization on GenImage, we maintain this identical UFD training configuration. For DRCT-2M, we follow the official protocol, training on 236k SDv1.4-generated images across 80 classes from MSCOCO~\cite{lin2014microsoft}.

\noindent\textbf{Testing Dataset}. The \textbf{UFD} provides fake images synthesized by 12 different GAN and DM generators. The \textbf{GenImage} dataset utilizes the ImageNet dataset as the source of real images and incorporates eight mainstream GAN and Diffusion generators to produce synthetic images. The \textbf{DRCT-2M} is composed of 13 comprehensive test sets that cover diverse diffusion architectures. Note that we exclude the three DR (Diffusion Reconstruction) subsets to focus on standard global synthesis. These 13 sets include multiple Stable Diffusion (SD) variants, latent consistency models (LCM), and ControlNet-based conditional generators~\cite{zhang2023adding}.

\setlength{\tabcolsep}{6pt}
\begin{table*}[t]
\centering
\caption{Performance comparison on GenImage in Acc. (\%). Each row corresponds to a detection method, and each column to a generative model. The last column reports the mean Acc across all models. Red and blue colors indicate the best and second-best results, respectively.}

\small
{
\begin{tabular}{lccccccccccc}
\toprule
Method & ADM & Midjourney & GLID & VQDM & SDv1.4 & SDv1.5 &  Wukong & BigGAN & Mean Acc \\
\midrule
CNNSpot~\cite{wang2020cnn} & 58.77 & 52.58& 55.00 & 53.67 & 51.58 & 51.83 & 50.25 & 79.54 & 56.65 \\
UnivFD~\cite{ojha2023towards} &66.87 & 56.13& 62.46 & 85.31 & 63.66 & 63.49 & 70.93 & 90.13& 69.87 \\
LGrad~\cite{tan2023learning} & 67.12 & 58.98& 62.98 & 70.08 & 68.41 & 69.00 & 65.48 & 81.32& 67.92 \\
NPR~\cite{tan2024rethinking} & 69.72 & {{78.00}} & 78.35 & 78.13 & 78.62 & 78.88 & 76.10 & 80.30 & 77.26 \\
FatFormer~\cite{liu2024forgery} & 79.49 & 55.52& 89.09 & 87.98 & 88.09 & 87.88 & 88.03 & {{98.67}} & 84.34 \\
VIB~\cite{zhang2025towards} & 72.20 & 58.50 & 69.75 & 87.15 & 67.50 & 67.50 & 74.00 & 88.55& 73.14 \\
ForgeLens~\cite{chen2025forgelens} & {{80.65}} & {{64.18}} & {{94.92}} & {{89.30}} & {{95.87}} & {{95.53}} & {{94.16}} & {{98.80}} & \color{blue}{\underline{89.18}} \\
\hline
\rowcolor{lightgray}{LTD(ours)} & {{90.38}} & 62.97 & {{97.63}} & {{92.53}} & {{97.23}} & {{97.24}} & {{96.33}} & 98.61& \color{red}{\textbf{91.62}} \\
\bottomrule
\end{tabular}
}
\label{tab:GenImage_acc}
\end{table*}
\setlength{\tabcolsep}{6pt}
\begin{table*}[t]
\centering
\caption{Performance comparison on GenImage in AP. Each row corresponds to a detection method, and each column to a generative model. The last column reports the mean AP across all models. Red and blue colors indicate the best and second-best results, respectively.}

\small
{
\begin{tabular}{lccccccccccc}
\toprule
Method & ADM & Midjourney & GLID & VQDM & SDv1.4 & SDv1.5 &  Wukong & BigGAN & Mean AP\\
\midrule
CNNSpot~\cite{wang2020cnn} & 71.10 & 55.94& 66.19 & 61.96 & 56.91 & 51.83 & 50.25 & 89.89 & 63.01 \\
UnivFD~\cite{ojha2023towards} & 86.81 & 74.00 & 83.81 & 96.53 & 86.14 & 85.84 & 91.07 & 98.13 & 87.79 \\
LGrad~\cite{tan2023learning} & 72.64 & 69.13& 76.28 & 74.36 & 71.85 & 72.51 & 68.03 & 88.83 & 74.20 \\
NPR~\cite{tan2024rethinking} & 74.64 & {{85.60}} & 85.72 & 81.79 & 84.03 & 84.59 & 80.47 & 88.40 & 83.16 \\
FatFormer~\cite{liu2024forgery} & 95.00 & 73.78& 97.72 & {{98.53}} & 98.37 & 98.14 & 98.62 & {{99.93}} & 95.01 \\
VIB~\cite{zhang2025towards} & 86.70 & 74.41& 88.66 & 96.22 & 85.13 & 83.99 & 90.15 & 97.29 & 87.82 \\
ForgeLens~\cite{chen2025forgelens} & 95.64 & {{83.72}} & {{99.03}} & 98.09 & {{99.24}} & {{99.19}} & {{99.19}} & {{99.97}} & \color{blue}{\underline{96.76}}  \\
\hline
\rowcolor{lightgray}{LTD(ours)} & {{98.74}} & 80.89 & {{99.74}} & {{99.18}} & {{99.72}} & {{99.61}} & {{99.58}} & 99.89& \color{red}{\textbf{97.17}} \\
\bottomrule
\end{tabular}
}
\label{tab:GenImage_ap}
\end{table*}

\noindent\textbf{Metrics}. Following previous work, we employ Accuracy (ACC) and Average Precision (AP) to evaluate the performance of detectors. Without prior distribution knowledge, we set a standard 0.5 decision threshold for Accuracy.

\noindent\textbf{Implementation Detail}. We employ the pre-trained CLIP ViT-L/14 model as the backbone for feature extraction and implement our framework on the Pytorch platform. During training, all input images are resized to $256\times 256$ resolution and then center-cropped to $224\times 224$. Random cropping and horizontal flipping are applied for data augmentation. All models are trained using the Adam optimizer~\cite{adam2014method} with a fixed learning rate of 5e-5 and a batch size of 256. Notably, our approach demonstrates exceptional training efficiency, converging in just 5 epochs on a NVIDIA RTX 4090 GPU.

\subsection{Comparing with the State-of-the-Art Methods}
\noindent\textbf{Comparisons on UFD}. Tables~\ref{tab:tab1} and~\ref{tab:tab2} present a comprehensive evaluation on the UFD benchmark~\cite{ojha2023towards}, where all baseline results are obtained using officially released models to ensure fair comparison. Notably, D$^3$~\cite{yang2025d} was trained on a broad spectrum of generators (2 GANs and 6 DMs). Other compared methods are primarily trained on ProGAN, covering either all 20 categories or at least 4 categories (\emph{car, cat, chair, horse}). In contrast, our LTD method requires only 2 categories (\emph{chair} and \emph{tvmonitor}) for training. As shown, our method outperforms all methods, which achieves the highest performance in mean Acc (\textbf{96.90\%}) and AP (\textbf{99.51\%}), surpassing recent state-of-the-art baselines ForgeLens~\cite{chen2025forgelens} and FatFormer~\cite{liu2024forgery} and  by \textbf{1.34\%} and \textbf{0.92\%} in mean Acc, respectively. Moreover, on the more challenging benchmark ADM that contains high-fidelity generation and complex noise scheduling, our method achieves \textbf{88.00\%} in Acc and \textbf{96.83\%} in AP. Such comparisons validate the superior effectiveness, generalizability, and training efficiency of our method. 

\noindent\textbf{Comparisons on DRCT-2M}. The quantitative results are illustrated in Table~\ref{tab:tab3} and Table~\ref{tab:tab4}. Overall, our LTD yields the best mean Acc of \textbf{99.54\%}, consistently exceeding all prior baselines. Compared to ForgeLens~\cite{chen2025forgelens} and UnivFD~\cite{ojha2023towards}, LTD acquires \textbf{+1.32\%} and \textbf{+10.20\%} gains in Acc. This demonstrates the generalization capability across both DM-based and controllable conditioned variants. Besides, as reported in Table~\ref{tab:tab4}, our LTD preserves perfect performance in mean AP (\textbf{99.99\%}), while other methods (\emph{e.g.}, AIDE \cite{yan2024sanity} and VIB~\cite{zhang2025towards}) exhibit severe performance drops when transferred from standard DMs (\emph{e.g.} LDM) to refined or accelerated versions (\emph{e.g.} LCM SDXL), suggesting limited robustness to generation dynamics.

\begin{figure}[t]
    \centering
    \includegraphics[width=\columnwidth]{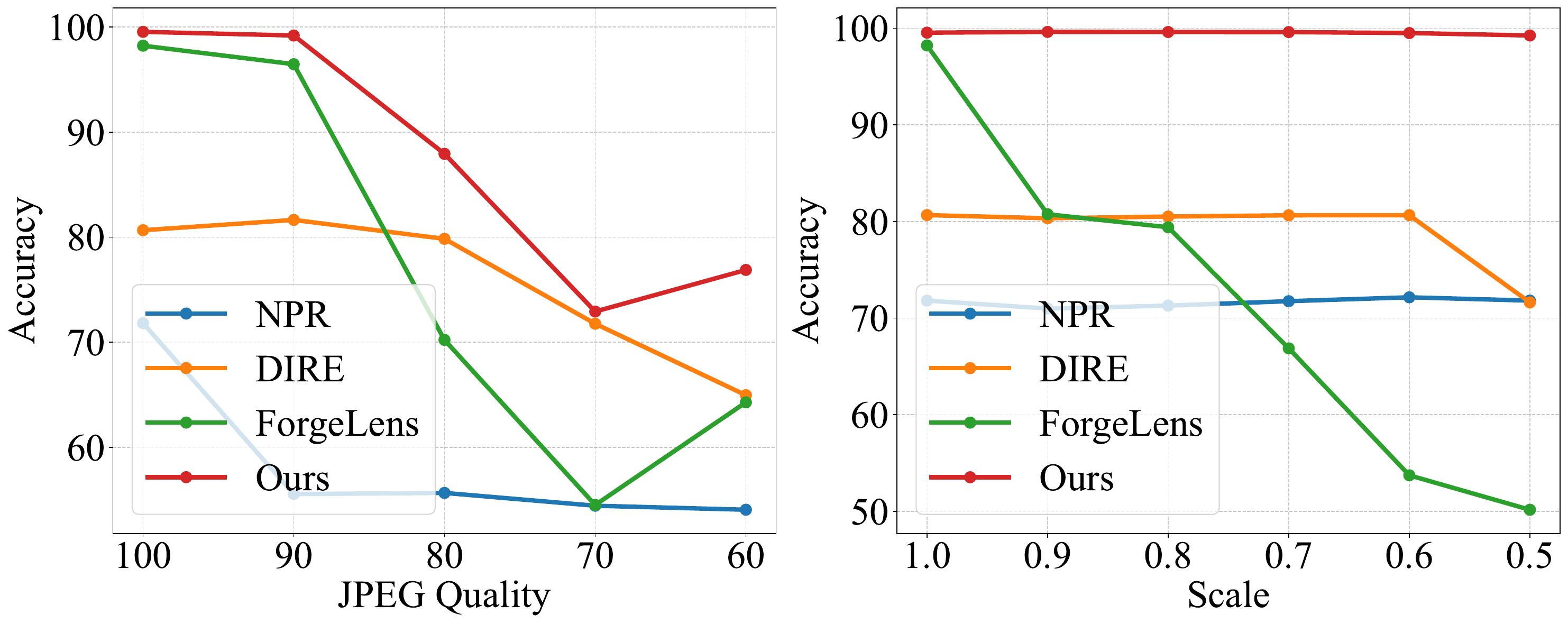} 
    \caption{Robust evaluation of our method and baselines under (a) JPEG compression with different quality factors and (b) downsampling with different downscaling factors. Our LTD validates superior robustness to different degradation types and levels.}
    \label{fig:fig5}
    \vspace{-0.5cm}
\end{figure}

\noindent\textbf{Comparisons on GenImage}. Table~\ref{tab:GenImage_acc} presents a comprehensive comparison between LTD and recent detection methods on the GenImage benchmark, with all methods trained on the same UFD training set to ensure a fair evaluation. Our LTD achieves the best performance across most generators and demonstrates a clear superiority in both metrics. As a result, LTD surpasses recent methods by significant margins on average. showcasing over 2.44\% improvement in mean Acc over the second-best method ForgeLens.

\noindent\textbf{Robustness Evaluation}. In real-world scenarios, images are frequently subject to post-processing operations such as compression and downsampling, particularly when shared through social media platforms. To evaluate our method, we conduct experiments on the \textbf{DRCT-2M} dataset focusing on two degradation types: (i) \textbf{JPEG compression} with quality factors ranging from 100 to 60, and (ii) \textbf{downsampling} with scale factors progressively reduced from 1.0 to 0.5. As shown in Figure~\ref{fig:fig5}, our LTD validates the superior robustness against recent state-of-the-art methods. In particular, with severe degradations, our method maintains a more stable detection performance across varying degradation levels, but other methods suffer from low accuracy.

\subsection{Ablation Study}
\textbf{Effect of LTD}. We conducted experiments to investigate the effect of the proposed LTD approach by implementing 4 model variants: \textbf{1)} The baseline model that performs the detection only using the selected raw mid-level features (Raw ML.); \textbf{2)} Introducing positional encoding (Pos. Enc.) in \textbf{1)}; \textbf{3)} Replacing \textbf{Raw ML.} with our LTD features, which performs detection only relying on the transition discrepancy; \textbf{4)} Introducing positional encoding in \textbf{3)}; \textbf{5)} The full model combines all components, which detects the real/fake images based on both raw mid-level and LTD features. As shown in Table~\ref{tab:module_ablation}, the baseline model yields the worst detection performance and the incorporation of positional encoding can significantly improve the accuracy as it can ensure position awareness regarding to the original hierarchy, thereby enabling the model to implicitly capture the underlying transition discrepancies across layers. We can also see that LTD is more effective than the raw features as it can characterize inter-layer variations while suppressing redundant irrelevant information. The combination of all components demonstrates the best performance.

\begin{table}[t]
\setlength{\tabcolsep}{6pt}
\renewcommand{\arraystretch}{0.9}
\centering
\caption{Ablation study on the layer transition consistency (LTD) detection. Raw ML. denotes raw mid-level features and Pos. Enc. denotes positional encoding, respectively.}

{
\resizebox{\linewidth}{!}{\begin{tabular}{ccc|ccc}
\toprule
Raw mid-level & LTD & Pos. Enc. & UFD & DRCT-2M & Mean Acc. \\
\midrule
\ding{51} & \ding{55} & \ding{55} & 84.92 & 92.75 & 88.84 \\
\ding{51} & \ding{55} & \ding{51} & 94.22 & 96.12 & 95.17 \\
\ding{55} & \ding{51} & \ding{55} & 86.42 & 93.50 & 89.96\\
\ding{55} & \ding{51} & \ding{51} & 92.43 & 94.01 & 93.22 \\
\ding{51} & \ding{51} & \ding{51} & \textbf{96.90} & \textbf{99.54} & \textbf{98.22} \\

\bottomrule
\end{tabular}
}}
\vspace{-0.3em}
\label{tab:module_ablation}
\end{table}

\noindent\textbf{Impact of Layer Selection}. In our method, before extracting the LTD features, we conduct dynamic layer-wise selection on mid-level features. To study its influence, we first analyze the discriminative power across different layer groups by partitioning all the 24 ViT layers into three non-overlapping segments: Shallow (0–7), Middle (8–15), and Deep (16–23). As shown in the first three rows of Table~\ref{tab:param_ablation}, the results confirm that separability between real and synthetic images is not uniformly distributed across layers. The Middle group demonstrates the strongest discriminative capability, while the Deep layers remain moderately effective. This observation aligns with our t-SNE visualization in Figure~\ref{fig:fig3}, where feature distinctions remain visible up to approximately layer 19, beyond which discriminative structure gradually diminishes. 

We further investigate the optimal starting point for mid-level feature selection by fixing the upper bound at layer 19 (guided by t-SNE analysis of feature discriminability) and varying the lower bound from 7 to 13. Performance improves as the starting layer moves deeper, peaking at layer 11 before saturating or slightly declining. This suggests that features spanning layers 11 to 19 capture the most distinctive inter-layer dynamics for distinguishing real from generated content. Finally, we examine how the number of selected layers affects performance. Varying the selection from 2 to 8 layers within the aforementioned optimal range (layers 11-19) reveals that performance consistently improves with more layers, reaching its peak at 5 layers before marginal degradation occurs due to redundant information. These findings demonstrate that a compact yet diverse subset of mid-level features provides the most discriminative representation while effectively mitigating overfitting.

\begin{table}[t]
\centering
\caption{Ablation study on the number of selected layers. Mean Acc is reported across different datasets.}
\small
{
\resizebox{\linewidth}{!}{
\begin{tabular}{c|ccccc}
\toprule
Layer Selection & UFD & DRCT-2M & GenImage & Mean Acc\\
\midrule
Fixed(0-7) & 62.87 & 82.12 & 45.70 & 63.56 \\
Fixed(8-15) & 95.78 & 98.98 & 90.38 & 95.50 \\
Fixed(16-24) & 92.62 & 97.57 & 79.78 & 89.99 \\
\midrule
Fixed(7-19) & 94.89 & 88.67 & 90.72 & 91.43\\
Fixed(9-19) & 95.02 & 95.75 & 91.34 & 94.04 \\
Fixed(11-19) & 96.26 & 97.44 &91.63 & 95.11 \\
Fiexed(13-19) & 95.57 & 95.52 & \textbf{92.06} & 94.38 \\
\midrule
2 Layers & 91.19 & 93.42 & 88.58 & 91.06 \\
4 Layers & 96.14 & 95.55 & 90.07 & 93.92 \\
5 Layers (\textbf{Final Setting}) & \textbf{96.90} & \textbf{99.54} & 91.62 & \textbf{96.02} \\
6 Layers & 96.19 & 97.76 & 91.76 & 95.24 \\
8 Layers & 96.80 & 97.56 & 91.38 & 95.25 \\
\bottomrule
\end{tabular}
}}
\label{tab:param_ablation}
\end{table}

\section{Conclusion}
This work establishes that synthetic images exhibit detectable layer transition discrepancies (LTD) in frozen mid-level ViT models, while real images maintain stable feature evolution across layers. Capitalizing on this finding, we propose the layer transition discrepancy (LTD) framework, which employs dynamic layer selection to identify the most discriminative mid-layer transitions and processes them through a dual-branch architecture. This design simultaneously captures holistic feature consistency and amplifies local inter-layer variations. Extensive experiments validate that our approach achieves state-of-the-art performance, demonstrating a significant accuracy improvement over baseline methods while maintaining superior robustness and generalization across diverse GAN and diffusion models with efficient implementation.

\section*{Acknowledgements}
This work was supported in part by the National Natural Science Foundation of China (62302141, 62331003, 62472138, 62272141).

{
    \small
    \bibliographystyle{ieeenat_fullname}
    \bibliography{main}
}

\clearpage

\appendix             

\setcounter{table}{0} 
\setcounter{figure}{0}

\section{Appendix}
In this supplementary material, we first provide a comprehensive introduction to the datasets used in our experiments (Sec.~\ref{sec:datasets}). Then, we conduct additional experiments on the Chameleon dataset~\cite{yan2024sanity} (Sec.~\ref{sec:expirements}). More ablation studies related to visualization analysis, different degradation factors, and alternative backbones are presented (Sec.~\ref{sec:ablation}).
\section{Datasets}
\label{sec:datasets}

\textbf{UnivFD}~\cite{ojha2023towards} contains synthetic images generated by a wide range of models including: 1) GAN-based methods.  ProGAN~\cite{karras2018progressive}, StyleGAN~\cite{karras2019style}, BigGAN~\cite{brock2018large}, CycleGAN~\cite{zhu2017unpaired}, StarGAN~\cite{choi2018stargan}, GauGAN~\cite{park2019semantic}, DeepFakes~\cite{rossler2019faceforensics++}; 2) Diffusion-based methods. ADM~\cite{dhariwal2021diffusion} trained on ImageNet, LDM~\cite{rombach2022high}, Glide~\cite{nichol2021glide}, DALL-E~\cite{ramesh2021zero}; and 3) The variants of some diffusion models. LDM with different noise refinement steps (\emph{e.g.}, 100 vs. 200, with or without classifier-free guidance) and Glide with multi-stage refinement (\emph{e.g.}, 100-27, 50-27, and 100-10). All samples are standardized to $256 \times 256$ resolution, with real counterparts sampled from the LAION~\cite{schuhmann2021laion} and ImageNet~\cite{russakovsky2015imagenet} datasets.

\noindent\textbf{DRCT-2M}~\cite{chen2024drct} is a large-scale benchmark comprising 2 million diverse synthetic images generated by state-of-the-art diffusion models (DMs), including Stable Diffusion variants (v1.4, v1.5, SDXL)~\cite{rombach2022high}, ControlNet-enhanced models (\emph{e.g.}, SDXL-Ctrl)~\cite{zhang2023adding}, and high-speed inference variants (\emph{e.g.}, SD-Turbo, LCM-SD). Images are sourced from community platforms (Civitai, Discord) to reflect real-world generation practices, capturing natural variations in prompt engineering, sampling configurations, and post-processing. Additionally, DRCT-2M contains a balanced subset of high-quality real photographs from established natural image collections to ensure comprehensive coverage of both authentic and synthetic content.

\noindent\textbf{GenImage}~\cite{zhu2023genimage} is specifically designed for detecting synthetic images produced by modern generative models, which involves 1,331,167 real images and 1,350,000 fake images, carefully balanced in terms of class distribution and image count. The training set consists of images generated by Stable Diffusion v1.4 ~\cite{rombach2022high}, paired with corresponding ImageNet ~\cite{russakovsky2015imagenet} labels to ensure semantic alignment between real and synthetic data. During evaluation, the detector is tested on a diverse range of generators, including both diffusion models (\emph{e.g.}, Stable Diffusion v1.5, GLIDE ~\cite{nichol2021glide}, ADM ~\cite{dhariwal2021diffusion}, VQDM ~\cite{gu2022vector}, Wukong~\cite{wukong2022} ) and GAN-based models (\emph{e.g.}, BigGAN ~\cite{brock2018large}), as well as commercial systems like Midjourney~\cite{midjourney2022}.

\noindent\textbf{Chameleon}~\cite{yan2024sanity} provides a high-quality and diverse collection of real and AI-generated images designed for evaluating detection robustness in realistic scenarios. It contains approximately 26K samples, with 14.8K real and 11.1K fake images, spanning four broad categories (human, animal, object, and scene), with resolutions ranging from 720P to 4K. The synthetic subset covers images generated from widely adopted text-to-image models such as Midjourney, Stable Diffusion (v1.4 and v1.5), DALLE-2, and various LoRA-based fine-tuned models. In Chameleon, the real subset is drawn from open-license platforms (\emph{e.g.}, Unsplash) to ensure distributional alignment.

\section{Comparisons on Chameleon}
\label{sec:expirements}

\begin{table*}[t]
\centering
\caption{Accuracy (Acc.) of different detectors (columns) on the Chameleon dataset when trained with different sources (rows). For each training dataset, the first row reports the overall classification accuracy, while the second row presents the class-wise accuracy split into ``fake/real'' for a more detailed breakdown.}

{
\resizebox{\linewidth}{!}{\begin{tabular}{cccccccc|c}
\toprule
\textbf{Training} & CNNSpot & LGrad & UnivFD & NPR & AIDE & D$^3$ & ForgeLens & LTD \\
\textbf{Dataset} & \cite{wang2020cnn} & \cite{tan2023learning} & \cite{ojha2023towards} & \cite{tan2024rethinking} & \cite{yan2024sanity} & \cite{esser2024scaling} & \cite{chen2025forgelens} & (ours)\\
\midrule
\multirow{2}{*}{\textbf{ProGAN}} & 57.31 & 59.41 & 57.36 & \textbf{59.83} & 58.38 & 56.63  & 57.71 & 58.27 \\
& 99.71/0.89 & 99.44/6.13 & 97.27/4.25 & 99.56/5.47 & 98.47/5.04 & 99.06/0.18 & 77.33/31.46 & 98.55/4.66 \\
\cmidrule(l){2-9}

\multirow{2}{*}{\textbf{SDXL-Turbo}} & 57.09 & 55.76 & 63.78 & 58.37 & 56.78 & 72.96 & 49.22 & \textbf{73.98} \\
& 100.00/0.00 & 95.08/3.45 & 44.45/83.11 & 95.69/8.71 & 98.71/0.98 & 63.30/85.81 & 5.55/92.64 & 67.63/82.43 \\
\cmidrule(l){2-9}

\multirow{2}{*}{\textbf{LCM-SDv1.5}} & 61.96 & 53.30 & 68.00 & 60.28 & 62.60 & 70.50 & 57.09 & \textbf{75.66} \\
& 96.16/16.45 & 73.81/26.00 & 54.94/81.06 & 87.98/23.41 & 86.66/30.68 & 72.99/67.18 & 100.00/0.00 &80.6/69.09 \\

\bottomrule
\end{tabular}
}}
\label{tab:tab1}
\end{table*}
To further evaluate the robustness of our method in detecting realistic AI-generated content encountered in the wild, we conduct experiments on the Chameleon dataset, which contains diverse and high-quality samples that are inherently challenging to human perception. The results are reported in Table~\ref{tab:tab1}. We observe that all the models trained on ProGAN perform poorly, as the distributional gap between ProGAN-generated samples and the advanced generators used in Chameleon is substantial, resulting in near-random detection performance. In contrast, when training on more recent generators such as SDXL-Turbo and LCM-SDv1.5, the models achieve markedly improved performance, emphasizing the necessity of contemporary training data for generalizing to advanced synthetic media. Notably, our method consistently exceeds all baselines in both training datasets, demonstrating superior adaptability across generative distributions and validating its strong potential for real-world media forensics applications.

\begin{figure}[t]
    \centering
    \includegraphics[width=\columnwidth]{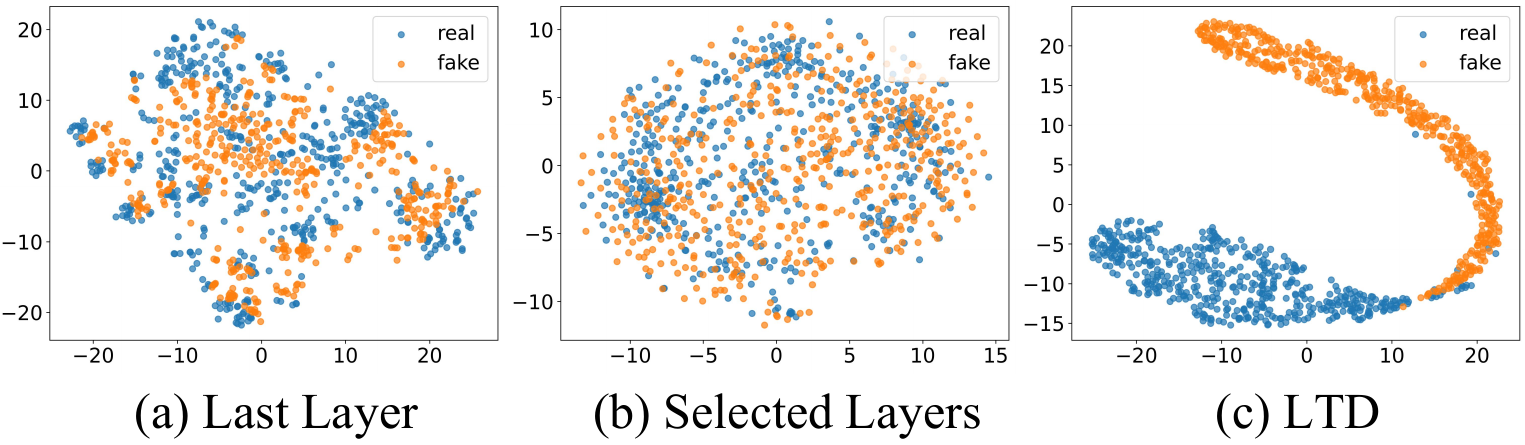} 
    \caption{t-SNE visualization of feature separability under different representations. Comparison among (a) the ViT last-layer features, (b) the concatenated selected mid-layer features, and (c) our LTD-enhanced representations. While both (a) and (b) exhibit limited separation between real and generated images, (c) shows clear and robust class separation, demonstrating the effectiveness of our LTD.}
    \label{fig:fig2}
\end{figure}
\section{Ablation Study}
\label{sec:ablation}

\noindent\textbf{Visualization of LTD}. To demonstrate the effectiveness of our approach, we visualize the feature distributions using t-SNE across three different settings: (i) the ViT last-layer representation (Last Layer), (ii) the concatenated features from our selected discriminative layers (Selected Layers), and (iii) our LTD representations. As shown in figure~\ref{fig:fig2}, both the last-layer and selected layers remain heavily overlapped, exhibiting limited separability between real and synthetic images. In contrast, our full model LTD produces a clearly separable embedding space, indicating that LTD effectively exposes inter-layer discrepancies and bring out discriminative artifacts that are otherwise not captured by the backbone alone.

\begin{figure}[t]
    \centering
    \includegraphics[width=\columnwidth]{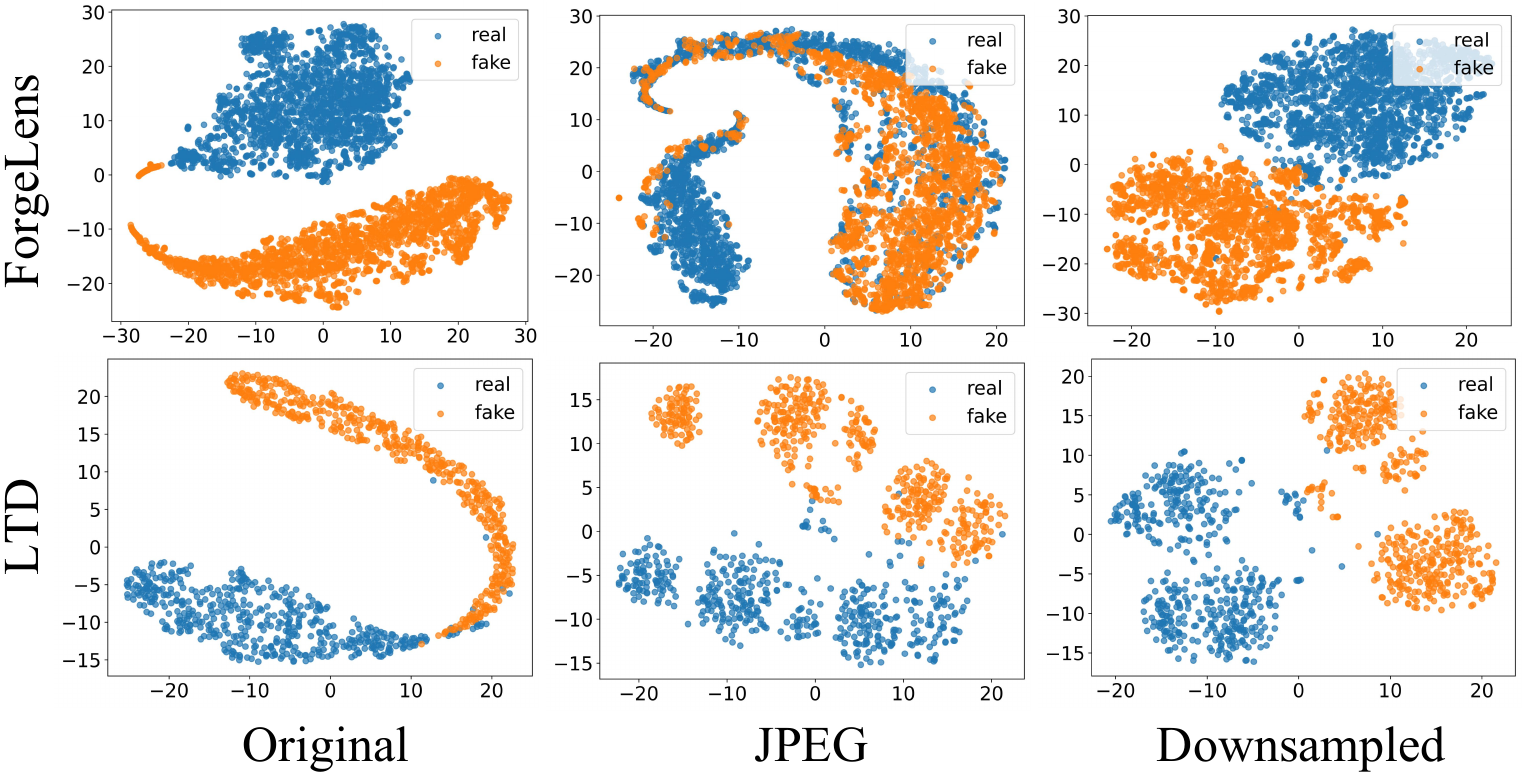} 
    \caption{t-SNE visualization of feature distributions on SD v1.4 generated images. Columns show features for original images, (b) JPEG-compressed images, and downsampled iamges. Under JPEG compression, ForgeLens suffers from severe cluster collapse. Notably, under downsampling, while ForgeLens retains visual separability, it exhibits a significant distribution shift that is mismatched with with the classifier (resulting in ~50\% Acc).}
    \label{fig:fig3}
\end{figure}

\begin{figure*}[t]
  \centering
  
  \begin{subfigure}[b]{\textwidth}
    \includegraphics[width=\textwidth]{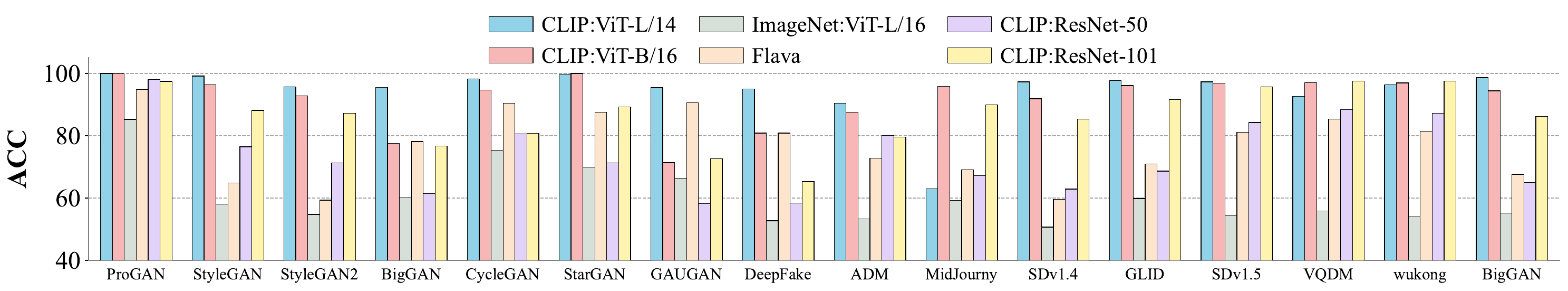}
  \end{subfigure}
  
  \vspace{3mm} 
  
  \begin{subfigure}[b]{\textwidth}
    \includegraphics[width=\textwidth]{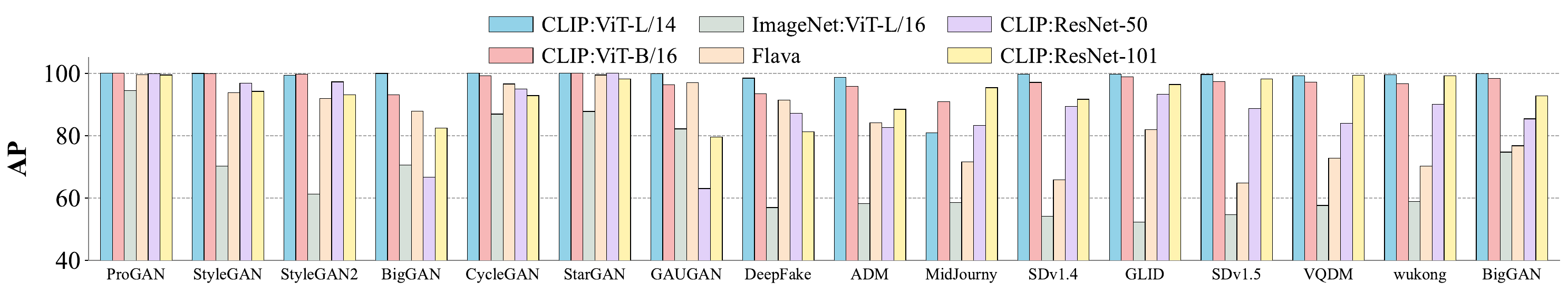}
  \end{subfigure}
  
  \caption{Backbone comparison across diverse generative models. Detection accuracy of four representative backbones, including CLIP ViT-L/14, CLIP ViT-B/16, ImageNet ViT-L/16, and Flava, evaluated over a wide range of GAN- and diffusion-based generators. The results highlight significant performance differences driven by pre-training strategy and model capacity.}
  \label{fig:backbone}
\end{figure*}

\noindent{\textbf{Robustness Against Realistic Degradations}}. To evaluate the robustness of our method under realistic degradations, we conduct perturbation experiments on the Chameleon, using LCM-SDv1.5 as the training set. Specifically, we consider three widely encountered distortions: JPEG compression with a quality factor of 80, downsampling with a scale factor of 0.5, and Gaussian blur with a kernel size of 3. Each degradation is applied individually to isolate its impact, and jointly to simulate more challenging compound distortions. As shown in Table~\ref{tab:tab2}, LTD maintains stable performance under JPEG compression and downsampling, with only marginal decreases of 1.20\% and 0.18\% in Acc., respectively. Blur introduces more substantial perturbations, resulting in performance drops of 6.52\% and 12.35\%. When all distortions are combined, the accuracy decreases by 4.77\% in Acc. and 7.37\% in AP. Despite these challenges, the overall robustness of LTD across various perturbations underscores its ability to capture coarse-grained, degradation-tolerant artifacts that remain reliable even when fine-grained frequency cues are severely disrupted.

\begin{table}[t]
\centering
\renewcommand{\arraystretch}{1.3}
\caption{Robustness analysis against image degradations on the Chameleon dataset. We evaluate the performance using average accuracy (Acc.) and average precision (AP). The values in parentheses denote the performance drop compared to the clean baseline.}
{
\resizebox{\linewidth}{!}{\begin{tabular}{ccccc}
\toprule
JPEG & Downsampling & Blur & Acc.(\%) & AP (\%)\\
\midrule
\ding{55} & \ding{55} & \ding{55}  & 75.66 & 78.00 \\
\ding{51} & \ding{55} & \ding{55} & 74.46 (1.20$\downarrow$)  & 77.16 (0.84$\downarrow$) \\
\ding{55} & \ding{51} & \ding{55} & 75.48 (0.18$\downarrow$)  & 77.41 (0.59$\downarrow$)\\
\ding{55} & \ding{55} & \ding{51} & 69.14 (6.52$\downarrow$) & 71.27 (6.73$\downarrow$) \\
\ding{51} & \ding{51} & \ding{51} & 70.89 (4.77$\downarrow$)  & 70.63 (7.37$\downarrow$) \\
\bottomrule
\end{tabular}
}}
\label{tab:tab2}
\end{table}

To more intuitively show the discriminative ability of our LTD in dealing with realistic degradations, we visualize the t-SNE in feature space, where the most recent method ForgeLens~\cite{chen2025forgelens} is employed as a reference. As shown in figure~\ref{fig:fig3}, in the ideal clean setting, both methods yield well-separated clusters between real and synthetic images. However, substantial differences emerge once degradations are applied. Under JPEG compression, LTD maintains a clear separation, while ForgeLens collapses into heavily overlapping clusters, showing almost no discriminative boundary. This behavior is consistent with the underlying mechanism of ForgeLens. ForgeLens is not a frequency-based detector, as it does not employ explicit spectral transformations or high-pass filtering. However, the method guides the frozen ViT to focus on forgery-specific local artifacts, which often reside in relatively high-frequency regions of the image. Consequently, when high-frequency content is suppressed by blur or corrupted by noise, these cues become less reliable, leading to the observed performance degradation.

For downsampling attacks, ForgeLens exhibits a significant performance discrepancy under downsampling attacks, where the detection accuracy drops to random guessing (as reported in the main manuscript) and the t-SNE visualization of the feature space reveals distinct, separable clusters. This mismatch indicates that downsampling does not fundamentally collapse the underlying feature distributions; instead, it disrupts the fine-grained local artifacts that ForgeLens relies on to form its decision boundary. As these high-frequency–sensitive cues are smoothed or removed, the classifier trained on them becomes invalid, even though coarser structural cues still induce separable feature clusters. 

By comparison, LTD maintains clear separability under both JPEG compression and downsampling, because its cross-layer consistency cues operate at a coarse granularity that is inherently resistant to such degradations.

\noindent{\textbf{Impact of Different Backbones}}. To examine the influence of different backbones on AI-generated images detection, we conduct experiments on several representative architectures, including CLIP-based ViTs, ResNets, ImageNet-pretrained ViT, and Flava~\cite{singh2022flava}. All backbones are trained on the 2-class training setting (\textit{chair}, \textit{tvmonitor}) described in the main paper, utilizing 72k ProGAN-generated images and real images from LSUN~\cite{yu2015lsun}. For the ResNet variants (CLIP:ResNet50~\cite{he2016resnet} and CLIP:ResNet101), since the feature dimensions vary across different stages, we apply Global Average Pooling to eliminate the spatial dimensions and align the features and project multi-stage features into a unified latent dimension. We summarize the performance across sixteen generators, spanning both GANs and DMs, as shown in Figure~\ref{fig:backbone}. the ImageNet-pretrained ViT-L/16 performs the worst, with accuracy dropping sharply across nearly all sources. This performance gap is primarily due to the significant difference in pre-training data scale; the limited volume of ImageNet-1K (~1.28M images) prevents the model from developing the robust sensitivity to generative artifacts that emerges from CLIP’s much larger dataset (~400M image-text pairs). Similarly, Flava, although its multimodal training, remains inferior to CLIP, further confirming that pre-training data scale is a important factor. CLIP’s extensive data provides a far superior prior for artifact sensitivity compared to the more limited datasets used for ImageNet or Flava. Furthermore, we observe a clear scaling effect across architecture, CLIP:ResNet-101 consistently outperforms CLIP:ResNet-50, and CLIP:ViT-L/14 exhibits greater stability than CLIP:ViT-B/16, which shows moderate but unstable performance on complex models like BigGAN and Stable Diffusion. These results prove that both increased model capacity and pre-training data scale are critical for developing a robust detector capable of handling the diverse artifacts of modern generative models.

\noindent\textbf{Impact of Weight sharing} To further investigate the efficacy of the weight-sharing mechanism within our dual-branch architecture, we conducted an additional ablation study comparing against a non-shared weight variant. In the non-shared setting, we assigned independent Transformer blocks to the selected raw feature branch and the LTD branch, respectively. The quantitative results demonstrate that the non-shared configuration leads to a consistent performance degradation across all evaluated benchmarks: on the UFD dataset , the accuracy dropped to 95.44\%, which is 1.46\% lower than the weight-sharing baseline; on the GenImage dataset , the performance decreased to 86.42\%, representing a margin of 3.20\% below the shared setting; and on the DRCT-2M dataset, the accuracy fell to 94.73\%, showing a substantial drop of 4.81\% compared to the proposed model. These results validate the critical role of weight sharing in our framework.

\begin{table}[t]
\centering
\caption{Ablation study on the selection endpoint (upper bound).}
\small
{
\resizebox{\linewidth}{!}{
\begin{tabular}{c|ccccc}
\toprule
Layer Selection & UFD & DRCT-2M & GenImage & Mean Acc\\
\midrule
Fixed(11-23) & 95.99 & 91.26 & 91.70 & 92.98 \\
Fixed(11-21) & 94.47 & 87.09 & 94.55 & 92.04 \\
Fixed(11-19) & 95.11 & 91.63 & 97.44 & 94.73 \\
Fixed(11-17) & 97.03 & 91.11 & 94.72 & 94.29 \\
\bottomrule
\end{tabular}
}}
\label{tab:endpoint}
\end{table}

\noindent\textbf{Impact of Weight sharing} To further investigate the efficacy of the weight-sharing mechanism within our dual-branch architecture, we conducted an additional ablation study comparing against a non-shared weight variant. In the non-shared setting, we assigned independent Transformer blocks to the selected raw feature branch and the LTD branch, respectively. The quantitative results demonstrate that the non-shared configuration leads to a consistent performance degradation across all evaluated benchmarks: on the UFD dataset , the accuracy dropped to 95.44\%, which is 1.46\% lower than the weight-sharing baseline; on the GenImage dataset , the performance decreased to 86.42\%, representing a margin of 3.20\% below the shared setting; and on the DRCT-2M dataset, the accuracy fell to 94.73\%, showing a substantial drop of 4.81\% compared to the proposed model. These results validate the critical role of weight sharing in our framework.

\noindent\textbf{Impact of Upper Bound} To further refine the optimal search space, we fixed the starting layer at Layer 11 (the optimal lower bound determined in our main text) and varied the endpoint from Layer 17 to Layer 23. As demonstrated in Table~\ref{tab:endpoint}, the detection performance consistently improves as the endpoint shifts from 23 towards 19. While extending the boundary to 17 yields the highest accuracy on the UFD dataset (97.03\%), it results in a significant performance drop on DRCT-2M (94.29\%) compared to the Layer 19 configuration. This indicates that a narrower selection at Layer 17 leads to overfitting and limited generalization across diverse datasets.


\end{document}